\documentclass[lettersize,journal]{IEEEtran}

\usepackage{amsmath,amsfonts,amsthm}

\usepackage{array}
\usepackage[caption=false,font=normalsize,labelfont=sf,textfont=sf]{subfig}
\usepackage{textcomp}
\usepackage{stfloats}
\usepackage{url}
\usepackage{verbatim}
\usepackage{graphicx}
\usepackage{cite}
\usepackage{color}
\usepackage{enumitem}
\usepackage[normalem]{ulem}
\usepackage{comment}
\usepackage{algorithm}
\usepackage{algpseudocode}

\usepackage{flushend}

\usepackage{hyperref}
\usepackage{mathrsfs}
\usepackage{booktabs}  
\usepackage{makecell}  
\usepackage{multirow}
\usepackage{pifont}
\usepackage[table]{xcolor}

\newcommand{\lld}[1]{{\color{olive} #1}}

\newcommand{\bs}[1]{\boldsymbol{#1}}
\newcommand{\cV}{\mathcal{V}}
\newcommand{\cM}{\mathcal{M}}
\newcommand{\tref}{\textup{ref}}

\theoremstyle{remark}
\newtheorem{remark}{Remark}

\begin{document}
\title{MARS-Dragonfly: Agile and Robust Flight Control of Modular Aerial Robot Systems}
\author{Rui Huang, Zhiqian Cai, Siyu Tang, Pengxuan Wei, Lidong Li, Xin Chen, Wenhan Cao, Zhenyu Zhang, Lin Zhao
\thanks{Rui Huang,  Siyu Tang, Pengxuan Wei, Lidong Li, Xin Chen, Wenhan Cao, Zhenyu Zhang, and Lin Zhao are with the Department of Electrical and Computer Engineering, National University of Singapore, Singapore 117583, Singapore (email: 
        {ruihuang@u.nus.edu, e1352616@u.nus.edu, luckystarwpx@163.com, li.ld@nus.edu.sg, e1499417@u.nus.edu, wenhan@nus.edu.sg, zhenyuzhang@u.nus.edu, elezhli@nus.edu.sg}).
        Zhiqian Cai is with the Engineering Design and Innovation Centre, National University of Singapore, Singapore 117583, Singapore (email: 
        {e1391152@u.nus.edu})}
}

\markboth{IEEE Journal Template}%
{A Sample Article Using IEEEtran.cls for IEEE Journals}


\maketitle

\begin{abstract}
Modular Aerial Robot Systems (MARS) comprise multiple drone units with reconfigurable connected formations, providing high adaptability to diverse mission scenarios, fault conditions, and payload capacities. However, existing control algorithms for MARS rely on simplified quasi-static models and rule-based allocation, which generate discontinuous and unbounded motor commands. This leads to attitude error accumulation as the number of drone units scales, ultimately causing severe oscillations during docking, separation, and waypoint tracking. To address these limitations, we first design a compact mechanical system that enables passive docking, detection-free passive locking, and magnetic-assisted separation using a single micro servo. Second, we introduce a force–torque–equivalent and polytope-constraint virtual quadrotor that explicitly models feasible wrench sets. Together, these abstractions capture the full MARS dynamics and enable existing quadrotor controllers to be applied across different configurations. We further optimize the yaw angle that maximizes control authority to enhance agility. Third, building on this abstraction, we design a two-stage predictive–allocation pipeline: a constrained predictive tracker computes virtual inputs while respecting force/torque bounds, and a dynamic allocator maps these inputs to individual modules with balanced objectives to produce smooth, trackable motor commands. Simulations of over 10 configurations and real-world experiments demonstrate stable docking, locking, and separation, as well as the effectiveness of the proposed modeling and control framework. To our knowledge, this work provides the first real-world demonstrations of MARS achieving agile flight and transport with 40° peak pitch motion, while maintaining an average position error of 0.0896 m across challenging tasks. The real-world experiment video can be accessed at: \url{https://youtu.be/yqjccrIpz5o}



\end{abstract}

\begin{IEEEkeywords}
 Cellular and Modular Robots, Aerial systems: Mechanics and control, Aerial systems: Applications.
\end{IEEEkeywords}
\section{Introduction}
\IEEEPARstart{C}{ollective} behavior is widely observed in various colonies of small animals and plays a pivotal role in activities such as predation and reproduction. As an example, male and female dragonflies cooperate in flight~\cite{kornova2024determines} and connect to lay eggs on submerged plant rhizomes~\cite{rodrigues2019egg}. 
Inspired by such natural forms of cooperation, modular aerial robot systems (MARS)\cite{garanger2025dodecacopter,oung2014distributed, gabrich2018flying, gandhi2020self, fixed_wing_carlson2022multi, su2023fault, huang2024adaptive, xu2025modular, wang2025versatile} offer an engineered platform in which multiple small drone units can flexibly adapt their geometry, payload capacity, and maneuverability through self-reconfiguration. By physically docking and separating in mid-air, MARS can form elongated structures for transportation, compact configurations for navigating cluttered spaces, or divide into smaller sub-groups for distributed tasks. These capabilities substantially expand the operational flexibility and efficiency of cooperative aerial systems.

Despite its great potential, the development of MARS still requires several demanding capabilities, including reliable docking~\cite{Modquad, ModQuad-Vi}, smooth in-flight separation~\cite{saldana2019design}, intelligent self-reconfiguration~\cite{saldana2017decentralized, litman2021vision, huang2025robustself_icra, huang2025transformars}, structural optimization~\cite{gabrich2021finding, su2024flight}, fault-tolerant control~\cite{li2024modular, huang2025mars-iros}, and motion planning~\cite {huang2025mars-iros,zhao2023versatile, Tokyo_chen2026hierarchical}. However, many of these tasks suffer from severe oscillations. We identify two key aspects for improving the agility and reliability of MARS: the \textbf{docking mechanism} and the \textbf{modeling and control framework}.

Regarding the docking mechanism, prior works~\cite{Modquad,saldana2019design} developed a purely magnetic docking system that enables passive attachment but requires large attitude maneuver for separation, leading to instability after detachment. 
Meanwhile, accurate position control is required for docking due to the small size of magnet (6.35~mm). 
Subsequent works~\cite{Tokyo_sugihara2023design,Tokyo_sugihara2024beatle} designed two servo motors to facilitate locking and separation. 
This design nevertheless increases system complexity and requires precise real-time docking detection. The above limitations motivate the development of a docking mechanism that achieves reliable attachment and separation while maintaining system simplicity.


In addition to the docking mechanism, the modeling and control framework is also critical for coordinated multi-drone operation. Prior work on MARS control remains limited and largely underdeveloped. Existing approaches typically rely on simplified models, Proportional–Integral–Derivative (PID) control, and rule-based force/torque allocation strategies~\cite{Modquad,ModQuad-Vi,saldana2019design}. These methods often require task-specific controllers for docking~\cite{Modquad,ModQuad-Vi}, separation~\cite{saldana2019design}, and yaw regulation~\cite{Modquad-dof}. Moreover, simplified models in prior work are typically valid only for small motions around hover, which can cause noticeable oscillations and lead to slow waypoint tracking. More critically, the discontinuous commands produced by error-based PID controllers and rule-based allocation are difficult for motors to track, causing attitude errors to accumulate as the number of robots increases and ultimately constraining both the admissible payload capacity and the system’s ability to achieve agile flight~\cite{Modquad}.

\begin{figure*}[t]
\centering
    \includegraphics[width = 18 cm]{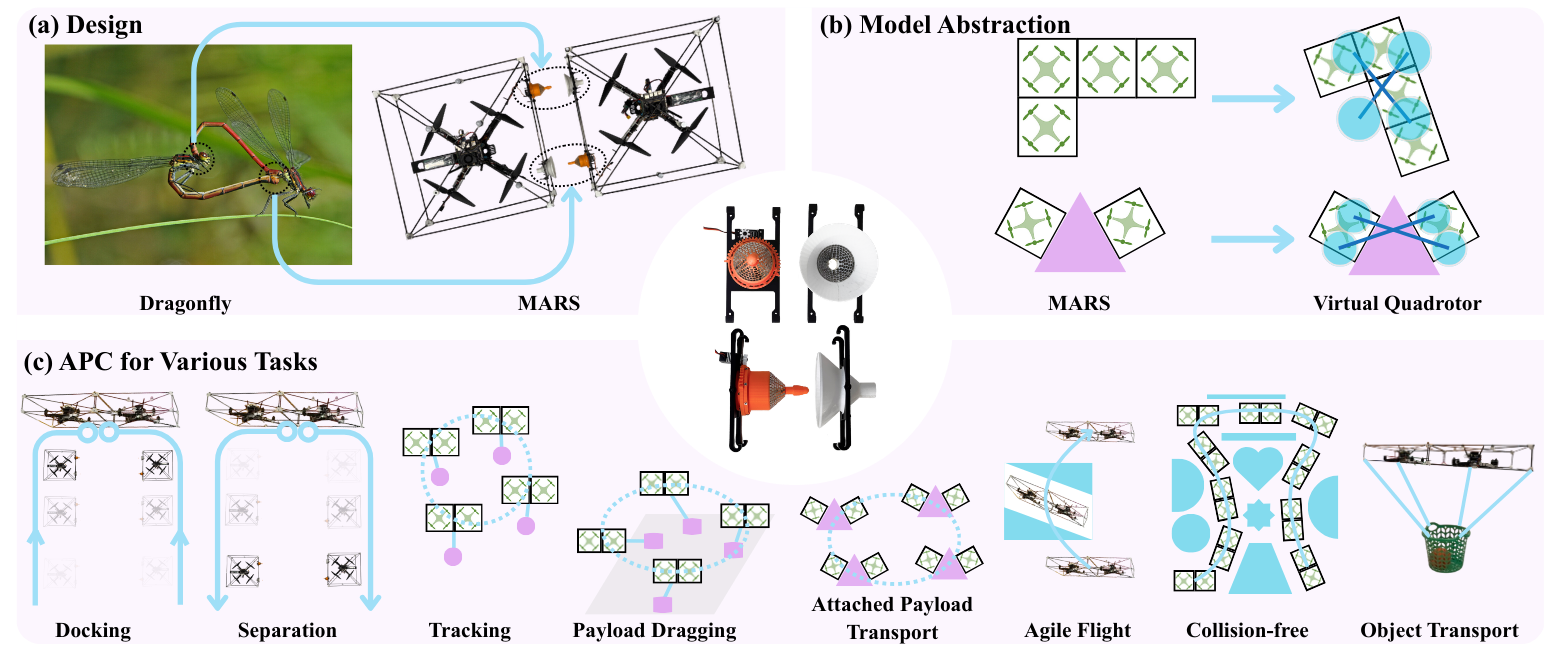}
    \vspace{-4mm}
\caption{Overview of the proposed MARS-Dragonfly framework.
 (a) Dragonfly-inspired MARS design.
 (b) A virtual quadrotor abstraction for arbitrary MARS configurations.
 (c) Enabling various MARS control tasks with a unified APC controller. }
\vspace{-6 mm}
\label{fig:prologue}
\end{figure*}

To address the above challenges, we develop a dragonfly-inspired docking–locking–separation mechanism for MARS and propose an Abstracted Predictive Control (APC) framework, as illustrated in Fig.~\ref{fig:prologue}. The framework employs a unified virtual-quadrotor model to capture transient flight dynamics and explicitly incorporates actuator-balanced constraints in the controller synthesis. The contributions of this work are summarized as follows: 
\begin{itemize}[nosep,noitemsep,left=0pt]

 \item A dragonfly-inspired docking–locking–separation unified mechanism is developed, which addresses the key limitations of prior MARS hardware. Unlike earlier platforms that implement docking and separation with independent mechanisms and provide no locking~\cite{Modquad,ModQuad-Vi,saldana2019design}, our design uses a self-aligning interface with magnetic assistance to tolerate significant approach misalignment, introduces a passive locking scheme that secures the connection without detecting the docking instant, and employs magnetic repulsion to achieve separation with minimal attitude change. This design reduces actuation and mechanical complexity compared with multi-link, dual-servo systems~\cite{Tokyo_sugihara2024beatle} while improving reliability across the full docking–locking–separation cycle.

 \item We develop an abstracted modeling framework for MARS that unifies the full system, including payload dynamics, into a single virtual quadrotor by preserving force and torque equivalence. Under this abstraction, we identify the optimal yaw angle to maximize control authority, thereby enhancing agility. This representation enables the direct application of a wide range of existing quadrotor control and planning algorithms, alleviating the maneuverability limitations imposed by prior quasi-static modeling approaches \cite{Modquad, ModQuad-Vi, Modquad-gripper}.

 \item We formulate a two-stage predictive–allocation optimization framework to address the scalability limitations of prior MARS. In the first stage, virtual control inputs are computed by minimizing deviations from desired state trajectories subject to the virtual quadrotor dynamics, while enforcing force and torque constraints to prevent actuator saturation. In the second stage, the resulting virtual control inputs are mapped to individual drone units in MARS through balanced dynamic control allocation. This design overcomes the oscillations caused by accumulated errors under discontinuous control commands in prior work~\cite{Modquad,ModQuad-Vi,Modquad-gripper}, enabling stable operation across arbitrary MARS configurations.

 \item We validate the proposed framework through extensive simulation and real-world experiments. In Gazebo and PX4 SITL simulations, we control MARS with different numbers of units across 11 configurations to demonstrate configuration-level generalization. We also conduct mid-air docking and separation experiments, where our mechanism shows markedly more stable and smoother behavior than prior systems~\cite{Modquad, ModQuad-Vi, Tokyo_sugihara2024beatle,saldana2019design}, achieving faster separation with negligible attitude disturbance compared to~\cite{saldana2019design}. Beyond docking and separation, we demonstrate previously unreported dynamic hardware capabilities, including agile flight and transport with $40^\circ$ pitch motion, collision-free navigation in cluttered environments, payload dragging, and sloshing payload transport.
\end{itemize}

The rest of this article is organized as follows. Section~\ref{sec:related works}  reviews related works. Section~\ref{sec:mechanism} presents the design of the compact docking-locking-separation mechanism. Sections~\ref{sec:model} and~\ref{sec:controller} develop the model abstraction algorithm and abstracted predictive controller, respectively. Section~\ref{sec:simulation} and~\ref{sec:real-world} present the simulation and real-world experiments, respectively. Section~\ref{sec:conclusion} concludes the paper.
\section{Related Works}
\label{sec:related works}
\begin{table*}[t]
\caption{MARS docking-locking-separation mechanisms}
\label{tab:comparison mechanisms}
\vspace{-2mm}
\centering
\scriptsize
\renewcommand{\arraystretch}{1.3}
\begin{tabular}{@{}lccclcc@{}}
\toprule
\makecell {\textbf{Mechanism}} 
& \makecell{\textbf{Mode}} 
&  \makecell{\textbf{ Docking Accuracy} \\ \textbf{Requirement}} 
& \textbf{Diagram} 
& \textbf{Docking or/and Separation Strategy}
& \makecell{\textbf{Actuators}  \\ (\textbf{a set})} 
& \makecell{\textbf{Load}} \\
\midrule

ModQuad ~\cite{Modquad} 
& Docking-only  
& \makecell{ High \\($\propto$ magnet size)}
& \raisebox{-0.5\height}{\includegraphics[width=2.5cm]{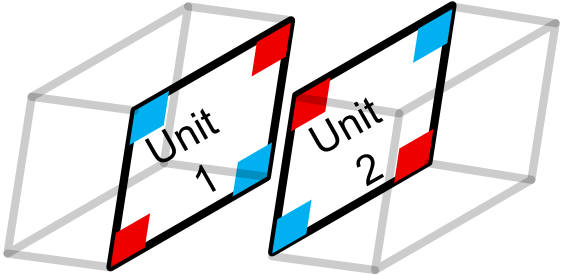}} 
& Magnetic attraction 
& 0 
& \makecell{Low} \\

ModQuad-Vi ~\cite{ModQuad-Vi} 
& Docking-only  
& \makecell{ High \\($\propto$ magnet size)}  
& \raisebox{-0.5\height}{\includegraphics[width=2.5cm]{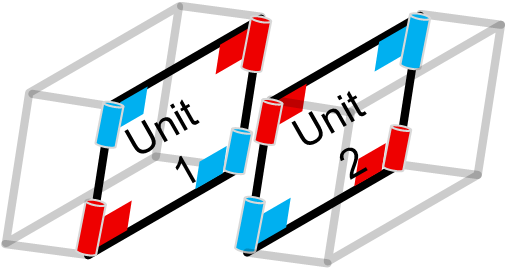}}  
& Magnetic attraction 
& 0 
& \makecell{Low} \\

Magnet matrix~\cite{saldana2019design} & Separation-only
& - 
& \raisebox{-0.5\height}{\includegraphics[width=2.5cm]{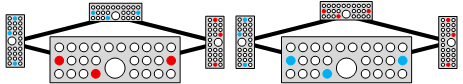}} 
& Separation torque due to attitude change
& 0 
& Low \\

Electromagnetic~\cite{zhang2024design}&  Separation-only
& -
& \raisebox{-0.5\height}{\includegraphics[width=2.5cm]{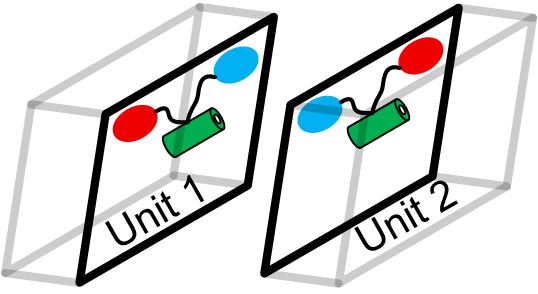}}  
& Switch the electromagnet on or off
& 4 
& \makecell{Medium} \\

\makecell[l]{BEATLE~\cite{Tokyo_sugihara2024beatle}} 
& Integrated 
& Medium 
& \raisebox{-0.5\height}{\includegraphics[width=3cm]{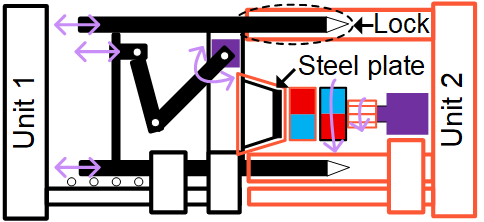}}  
& \makecell[l]{ {\textbf{Docking}: magnetic attraction} \\ {\textbf{Locking}: servo-actuated slider-crank} \\
{with stick insertion.} \\ 
{\textbf{Separation}:  a servo reverses the polarity }\\ 
{of the magnet, and a servo-driven slider-}\\ 
{crank mechanism retracts locking stick.}\\
} 
& 2 & \makecell{High} \\

\textbf{Our Design} 
& \textbf{Integrated}
& \makecell{\textbf{Low}}
& \raisebox{-0.5\height}{\includegraphics[width=3cm]{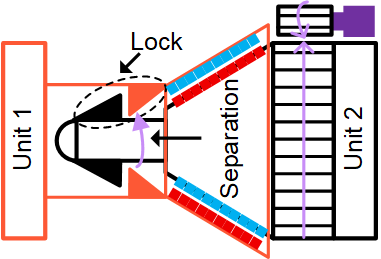}} 
&  \makecell[l]{
{\textbf{Docking}: magnetic attraction} \\ 
{\textbf{Locking}: asymmetric magnet-actuated }\\
{twist-lock for passive mechanical interlocking}\\
{\textbf{Separation}: a servo-driven magnetically }\\
{repulsive separation}
}
& \textbf{1} 
& \textbf{High} \\
\bottomrule
\end{tabular}
\footnotesize\textit{Red/blue: magnet poles (N/S); purple: servo motors.}
\label{tab:docking mechanism comparison}
\vspace{-6mm}
\end{table*}
\subsection{MARS Docking-Locking-Separation Mechanisms}
We review the mechanisms separately for docking-only, separation-only, and integrated docking-separation mechanisms. An overall summary is presented in Table~\ref{tab:comparison mechanisms}.

\subsubsection{Docking-only}
The works in \cite{Modquad, ModQuad-Vi} design docking mechanisms that leverage magnetic attraction, as illustrated in the Diagram column of Table~\ref{tab:docking mechanism comparison}, rows 1 and 2.
However, since the magnet contact area ($6.35 ~\text{mm}\times 6.35 ~\text{mm}$) is relatively small in their design, an accurate pose control is required for successful docking. Besides, this docking mechanism does not include a locking feature, and the tangential holding force is insufficient for stable lateral support.
Consequently, the system has the latent risk of unintended detachment caused by payloads, disturbances, or attitude maneuver.

\subsubsection{Separation-only}
The literature \cite{saldana2019design} and \cite{zhang2024design} designed separation mechanisms. 
In particular, \cite{saldana2019design} miniaturized the magnets and arranged them in a $3 \times 10$ matrix (cf. Table.~\ref{tab:docking mechanism comparison} row 3). 
Then the separation is achieved by weak tangential magnetic force and attitude-maneuver-induced torque, which results in the attitude angle change exceeding $25^\circ$. 
Besides, experimental results of \cite{saldana2019design} showed substantial position oscillations after separation, which will disrupt the flight performance. 
The separation in \cite{zhang2024design} employs circular electromagnetic chunks (cf. Table.~\ref{tab:docking mechanism comparison} row 4),
which requires eight power and control channels (two per edge), leading to increased energy consumption and control complexity. 
This issue would be a major constraint for power-limited aerial platforms.
Both \cite{saldana2019design} and \cite{zhang2024design} also lack an locking mechanism, which poses a potential risk of unintended detachment.

\subsubsection{Integrated Docking–Locking–Separation Mechanism}
Recent work in \cite{Tokyo_sugihara2024beatle} proposed an integrated docking-locking-separation mechanism for modular fully actuated tilt-rotor quadrotors, as illustrated in the fifth row of Table~\ref{tab:comparison mechanisms}.
In this design, a cone-shaped drogue on the female module (Unit 2) passively guides the male connector (Unit 1) during docking. 
A slider-crank mechanism then actuates locking sticks from the male side into reinforced sockets on the female module, mechanically locking the connection. 
For separation, the sticks are retracted using a servo-driven slider-crank assembly; meanwhile a servo-actuated switchable magnet reduces the magnet force, enabling detachment through flight-induced motion. However, several limitations persist: 
\textit{a) Weight:}
This mechanism is heavier than all prior designs, as it incorporates male and female structures, two locking sticks, two servo motors, and a slider-crank assembly. Although the components are densely integrated, the cumulative mass of the mechanism reduces agility and increases energy consumption.
\textit{b) Docking State Detection:}
Real-time detection of successful docking prior to locking is necessary, 
since premature activation of the locking mechanism may result in collisions with neighboring quadrotors. 
Moreover, extra sensors are needed for detection.
\textit{c) Mechanical Redundancy:}
The locking relies on servo-driven engagement, without secondary mechanical restraint. As such, the system is vulnerable to actuator failure and power loss. 
\textit{d) Docking Accuracy:}
Although the mechanism has a 70~mm diameter, which is larger than prior designs\cite{Modquad, ModQuad-Vi}, 
its flat-headed male connector (steel plate) requires stringent control precision, cf. Table.~\ref{tab:docking mechanism comparison} row 5. 
This requirement arises from the absence of a docking guidance structure, which makes the system susceptible to misalignment.

In summary, the above review and analysis of existing designs motivate the development of a new integrated docking-locking-separation mechanism that ensures light weight, reliable docking, detection-free locking, and safe separation.

\begin{table*}[t]
\caption{MARS model and control}
\label{tab:comparison model and control}
\vspace{-2mm}
\centering
\scriptsize
\renewcommand{\arraystretch}{1.3}
\begin{tabular}{@{}lcclc@{}}
\toprule
\makecell {\textbf{Method}} 
& \makecell{\textbf{Model}}
&  \makecell{\textbf{Controller}} 
& \textbf{Task}
& \makecell{\textbf{Control Performance}} \\
\midrule
ModQuad~\cite{Modquad} & Simplified model  & PID + Rule-based control allocation &  Docking, hovering & Severe oscillation \\

ModQuad-Vi~\cite{ModQuad-Vi} & Simplified model & PID + Rule-based control allocation &   Docking, hovering & Severe oscillation\\

ModQuad-DoF~\cite{Modquad-dof} & Simplified model & PID + Rule-based control allocation & Yaw control, hovering & Medium oscillation \\

Modquad-Gripper~\cite{Modquad-gripper} & Simplified model & PID + Rule-based control allocation & \makecell[l]{Waypoint tracking\\Object transport}  & Medium oscillation \\

Ref.~\cite{saldana2019design} & Simplified model & Torque peak generation &  Separation, hovering  & Severe oscillation \\

\rowcolor{gray!8}
\textbf{Ours} 
&  \makecell{\textbf{Virtual quadrotor }\\ \textbf{(Nonlinear model)}}
& \makecell{\textbf{Abstracted predictive control} \\+\\ \textbf{Balanced dynamic control allocation}}
& \makecell[l]{
{Docking, Separation}\\
{Trajectory tracking}\\
{Agile flight}\\
{Collision-free navigation}\\
{Object dragging}\\
{Agile transport}} 
& \textbf{Smooth and Stable} \\
\bottomrule
\end{tabular}
\vspace{-6mm}
\end{table*}
\subsection{MARS Model and Control}
The modeling of MARS in \cite{Modquad, ModQuad-Vi, Modquad-dof} used the total control effectiveness formulation, that is, the total thrust and torque are simply regarded as the sums of the thrust and torque generated by each unit. This simplified model assumes small perturbations around a hovering equilibrium in mid-air.
After the modeling step, most previous works adopted PID controllers to perform tasks such as docking, hovering, and yaw control.
Once the PID computes the desired total thrust and moments, rule-based strategies allocated the commanded forces to each rotor.
As for the task of separation, \cite{saldana2019design} tilted the quadrotor to induce torque peaks to break the magnetic linkage between modules.
Existing works have effectively modeled and controlled the MARS, but there is still room for further improvement:
\subsubsection{Modeling}
The control effectiveness formulation used in~\cite{Modquad, ModQuad-Vi, Modquad-dof, Modquad-gripper, saldana2019design} considers only a simplified model near hovering and also not fully exploit the geometric configuration of MARS to generate the maximum achievable moments.
In contrast, we first employ the dynamic virtual quadrotor model in~\eqref{eq:APC_model}. As formulated in \eqref{eq:op_orientation} or \eqref{eq:flex_op_plbm_obj} in our later sections, a larger control moment can be achieved by optimizing the MARS orientation through a rotation about the $z$-axis.
In addition, existing modeling methods usually do not explicitly incorporate the force limits of each rotor, whereas our formulation, as in \eqref{eq:flex_op_plbm}, naturally accommodates these constraints.
\begin{figure}[!t]
\centering
    \includegraphics[width = 8.8 cm]{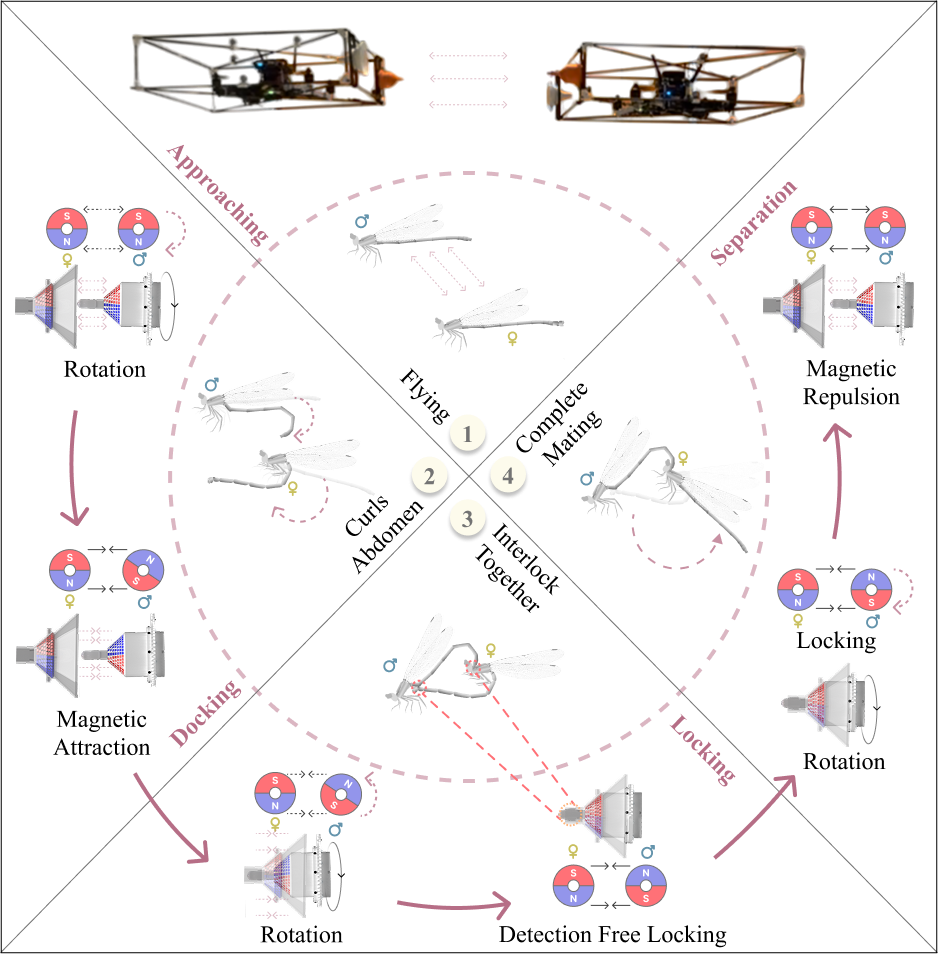}
\vspace{-8mm}
\caption{Docking, locking and separation sequence inspired by dragonfly mating behavior. The process replicates sequential biological actions, including flying \textcircled{\scriptsize 1}, curling the abdomen \textcircled{\scriptsize 2}, interlocking together \textcircled{\scriptsize 3}, and completing mating \textcircled{\scriptsize 4}, through staged magnetic attraction docking, passive locking, and controlled magnetic repulsion for active separation.}
\vspace{-6 mm}
\label{fig:dragonfly process}
\end{figure}
\subsubsection{Controller}
Existing works~\cite{Modquad, ModQuad-Vi, Modquad-dof, Modquad-gripper} employed error-based PID controllers. Although PID control is simple and practical, its response delay tends to accumulate and eventually causes oscillations as the number of MARS units increases. In contrast, the abstracted predictive controller in our framework incorporates system dynamics and future trajectory predictions and is capable of handling constrained control problems, thereby generating trackable commands for the virtual quadrotor and effectively mitigating the accumulated response delay in multi-unit MARS.

\subsubsection{Control Allocation}
In previous works \cite{Modquad, ModQuad-Vi, Modquad-dof,  Modquad-gripper, saldana2019design}, the total thrust and moment are obtained by distributing forces to individual rotors according to pre-defined rules. This approach produces discontinuous motor commands, which are difficult for the actuators to track, lead to accumulated attitude errors as the number of robots increases, and ultimately result in oscillatory behavior.
In contrast, as formulated in \eqref{eq:op_allocate}, our allocation scheme is cast as an optimization problem that enforces balanced constraints across all rotors and is solved at every control step, thereby balancing lift distribution and generating smooth, trackable motor commands.

To conclude, a dynamic model with proper constraints and a control allocation scheme that produces continuous commands is required. The existing modeling and control methods for MARS are summarized in Table~\ref{tab:comparison model and control}.
\section{Docking-Locking-Separation Mechanism}
\label{sec:mechanism}
The mating process of dragonflies consists of several stages: approach, docking, locking, and separation as shown in Fig.~\ref{fig:dragonfly process}. 
Meanwhile, the physiological structure of dragonflies ensures the stability and reliability of their mating process.
Inspired by these stages and the corresponding biological mechanisms, we designed our docking mechanism for the MARS.
\begin{figure*}[!t]
    \centering
    \includegraphics[width=\textwidth]{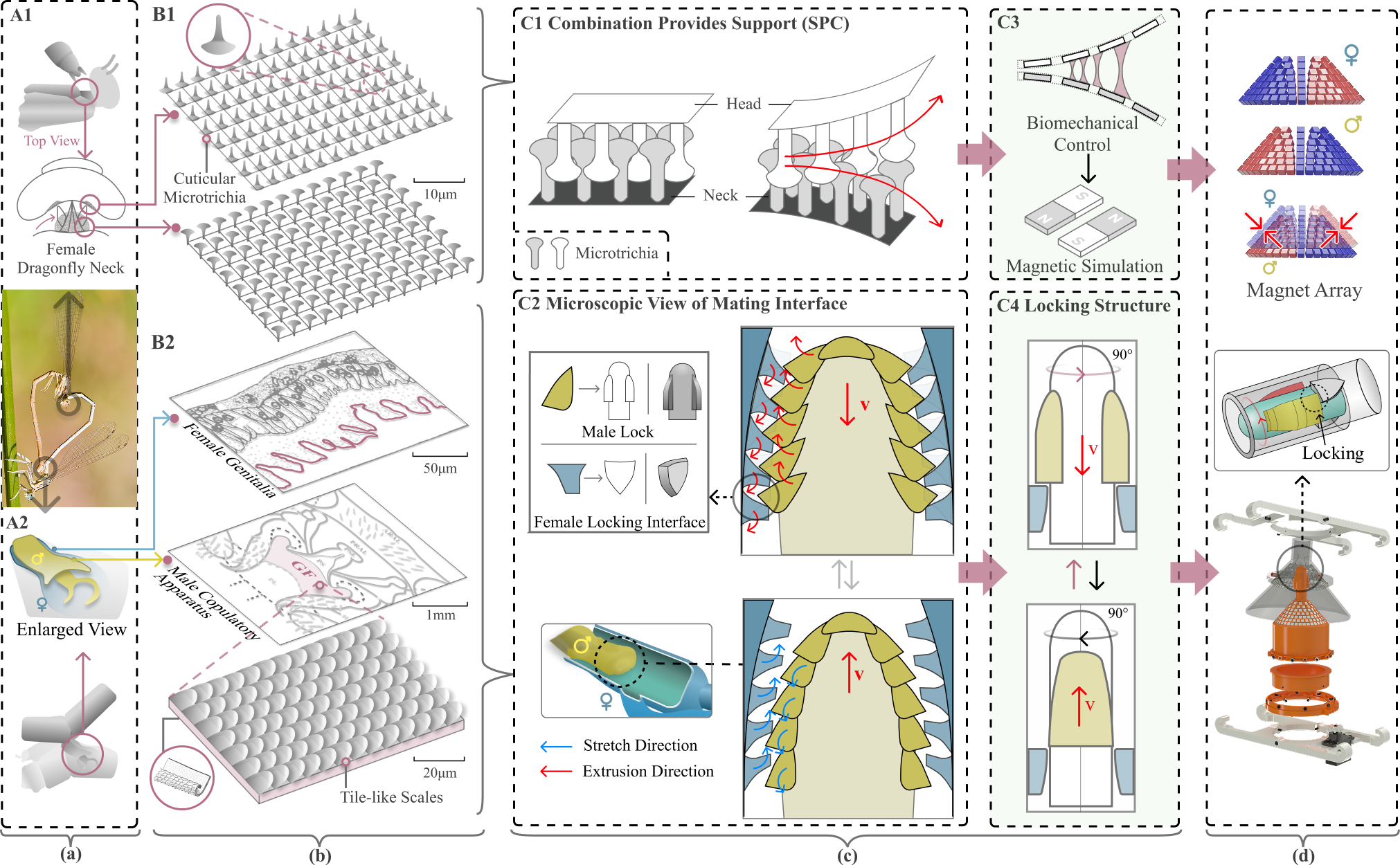}
    \vspace{-8mm}
    \caption{Microstructural connections in dragonfly mating that inspire the design of macroscopic mechanical structures. (a) The connection of dragonflies includes two docking parts. A1: the male grasps the female's neck. A2: the secondary genitalia of the male engages with the terminal end of the female's abdomen for copulation. (b) Microscopic structure of two docking parts. B1: The dense protrusions provide opposing frictional surfaces to support the neck during mating. B2: The microstructure of the female genital cuticle and tile-like scales of the male copulatory apparatus.
    (c) C1: To mimic the biomechanical control, we employ magnets to simulate a stable connection. C2: The epidermal structure (blue) and tile-like scale surface (yellow) provide enhanced stability. Inspired by this, we design a locking interface with a similar stabilizing effect. (d) The docking mechanism design with magnet array and locking interface.}
    \label{fig:micro-structure}
    \vspace{-6mm}
\end{figure*}
\subsection{Dragonfly-Inspired Design Overview}
\subsubsection{Docking}
The pairing begins as the male curves its abdomen to grasp the female’s prothorax. The female then bends her abdomen toward the male’s secondary genitalia to form the mating posture \cite{willkommen2015functional}.
In our bio-inspired design, as shown in Fig.~\ref{fig:dragonfly process}(2) and Fig.~\ref{fig:micro-structure}(a), this stage is mimicked as two robots approach each other, while the male and female docking interfaces rotate in advance to align their magnetic poles for attraction and engagement.
\subsubsection{Locking}
As illustrated in Fig.~\ref{fig:dragonfly process}(3), the two docking parts engage once the male and female dragonflies achieve initial aerial contact. 
Specifically, after docking, the dense protrusions on the contact surfaces between male and female dragonflies, as shown in Fig.~\ref{fig:micro-structure}(b1) and (c1), induce frictional and attractive forces. 
Moreover, dragonflies have barb-like structures in the docking regions to enhance locking, cf. Fig.~\ref{fig:micro-structure}(b2) and (c2).
In our mechanism, we use magnet arrays to replicate the attractive force of the dense protrusions, as illustrated in Fig.~\ref{fig:micro-structure}(c3), which shows the principle from biomechanical control to magnetic simulation. 
We further design the male and female locking interfaces to emulate the dragonfly’s barb-like structures, as shown in the top-right of Fig.~\ref{fig:micro-structure}(c2).
Owing to the self-locking arrangement of the magnet array [cf. Fig.~\ref{fig:micro-structure}(d)], the interface automatically rotates immediately after docking, which is driven by the tangential magnetic forces, thereby aligning the mechanism at the position of strongest magnetic attraction.
At this position, the barb-shaped locking interface is fully engaged, ensuring mechanical locking.


\subsubsection{Separation}
After mating, the female dragonfly rotates her abdomen to detach from the male, and the male curls his abdomen to release the female, as depicted in Fig.~\ref{fig:dragonfly process}(4).
In our mechanism, a micro motor rotates the male docking component to switch its magnetic polarity, generating a repulsive force to achieve active separation. 
Meanwhile, the barb-shaped head component is also rotated to the release position.

\subsection{Mechanism Details}

\begin{figure*}[!t]
\centering
    \includegraphics[width = 17.5 cm]{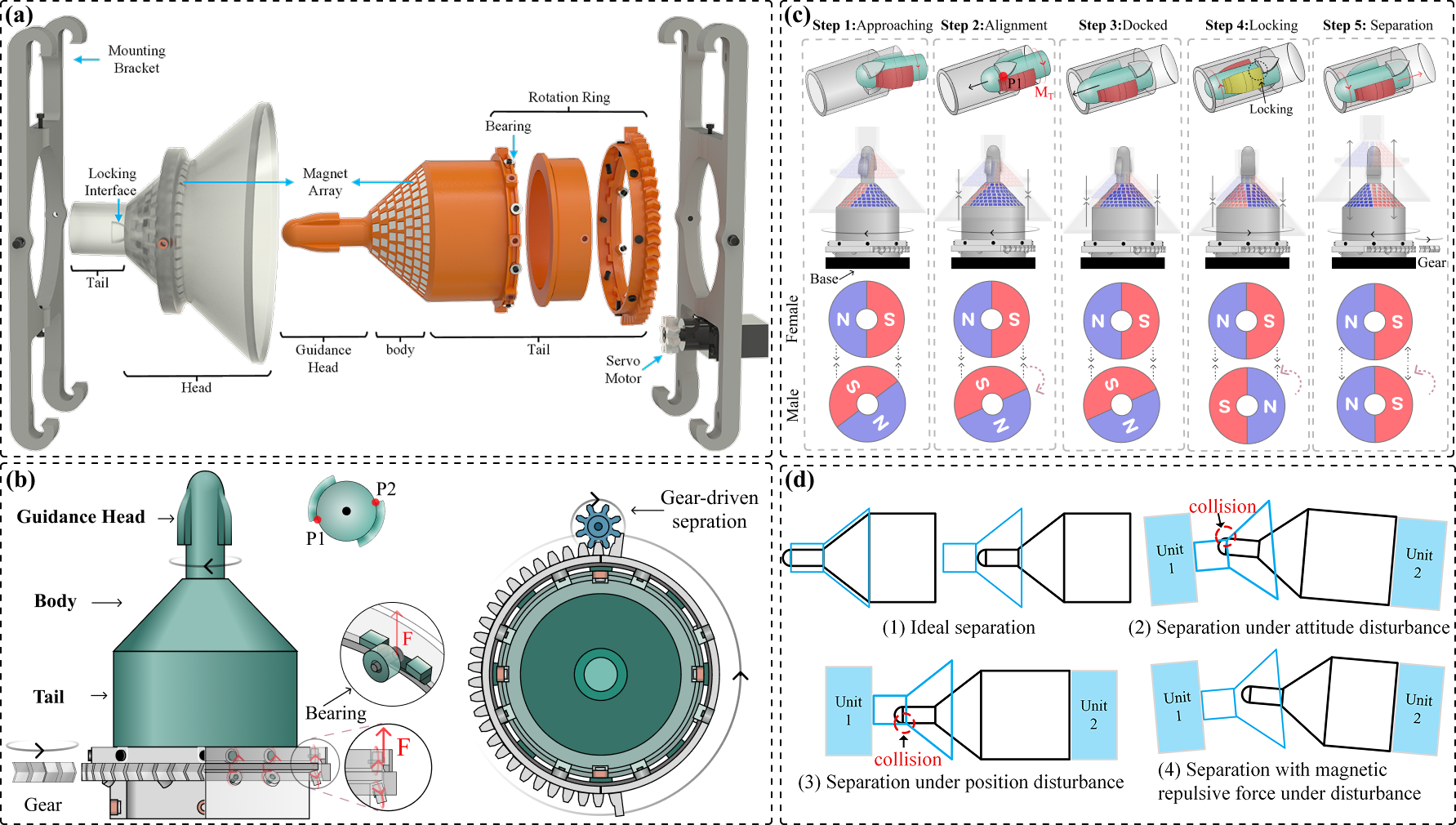}
\vspace{-3mm}
\caption{Docking and separation mechanism detail.  
(a) Exploded view of the mechanism. (b) Front and transparent views of the docking structure. Bottom view illustrating the circular gear arrangement that enables controlled separation. (c) Sequential process of  approaching, alignment, docking, passive locking, and gear-actuated separation. (d) Illustration of separation failure modes under different disturbances.}
\label{fig:mechanism_detail}
\vspace{-6mm}
\end{figure*}

In this subsection, we translate the dragonfly-inspired structure into a concrete mechanism design.
The docking mechanism consists of male and female components mounted on the carbon fiber frame of the MARS, as illustrated in Figs \ref{fig:mechanism_detail}(a). Further details are provided below.



\subsubsection{Male Docking Component}
The male docking mechanism consists of a docking guidance head, body, tail (including a rotation ring), and mounting bracket, as shown in orange part of Fig.~\ref{fig:mechanism_detail}(a).
To accommodate larger alignment errors, the male docking body is fitted with a cylindrical extension with an arc-shaped tip, i.e., the guidance head.
The guidance head carries barb-like structures used for locking. 
The male docking body is hollow and conical, with magnet arrays embedded in its outer surface, as depicted in Fig.\ref{fig:mechanism_detail}(a).
Following the body, the tail contains a bearing and a gear ring for rotation.
The rotation is driven by a micro motor mounted on the bracket.
In particular, during docking preparation, the motor rotates the body to the magnetically attractive position relative to the female docking component; during locking, the magnetic forces drive the body to rotate automatically to the position of maximum attraction; for separation, the motor rotates the body to the magnetically repulsive position.
Across these three stages, the barbs on the guidance head are rotated respectively to the easy-entry position, the locked position, and the easy-exit position.





\subsubsection{Female Docking Component}
The female docking component [grey part in Fig. \ref{fig:mechanism_detail} (a)] comprises three main parts: the docking head, docking tail, and mounting bracket. The head features a funnel-shaped geometry, while the tail forms a hollow cylindrical column. 
Similar to the male component, magnet arrays are embedded along the inner surface of the head, with their polarities arranged as shown in Fig.\ref{fig:mechanism_detail}(c).
The locking interface in the female tail mates with the male guidance head barb for easy entry, locking, and easy exit.
\subsection{Docking–Locking–Separation Process}
\subsubsection{Docking}
As shown in Fig.~\ref{fig:mechanism_detail}(c) Step 1, when the convex male docking component approaches the concave female docking component, 
due to the specific magnet arrangement, the magnet array on the male side and its counterpart on the female side produce an attractive force, helping guide the two quadrotors into docking.
Normally, the female and the male locking interfaces (the barb-shaped structures) do not collide during the docking process; 
but even if they do, both the female and the male locking interfaces are designed with a self-aligning geometry. 
As shown in Fig.\ref{fig:mechanism_detail}(c), 
due to the specially designed chamfered surfaces, contact between the two components generates a moment $M_T$ that guides the mechanisms into alignment.

\subsubsection{Locking}
A key advantage of the proposed mechanism is its ability to achieve detection-free passive locking. 
As shown in Steps 3 to 4 of Fig.~\ref{fig:mechanism_detail}(c), 
thanks to the arrangement of the magnet arrays on the male and female components, the torque generated by the tangential magnetic force rotates the male component into the locking position.
Finally, the male locking structure engages with the female interface, completing the passive locking.


\subsubsection{Separation}
A servo motor embedded in the male docking component drives a rotation ring via a herringbone gear mechanism, rotating it to the designated separation angle. The detailed gear configuration is shown in Figs.~\ref{fig:mechanism_detail}(a) and (b). As the male component rotates, the polarity of its magnet array is reversed, generating a repulsive magnetic force. 
We emphasis that the repulsive magnetic force is crucial in separation.
On one hand, various disturbances will affect the quadrotor's attitude and position. 
On the other hand, underactuated quadrotors must tilt their attitude to achieve position motion. 
As a result, collisions may occur during separation through purely quadrotor movement, which may cause failure, as illustrated in Fig.\ref{fig:mechanism_detail}(d). This type of separation failure is occasionally observed in our experiments; see Fig. \ref{fig:failed_flight_separation} for details.
Fortunately, the magnetic repulsion can open a small gap between the male and female components at the moment of separation, enabling smoother separation with a higher success rate.

Overall, our proposed design exhibits the following key features: 1) integrated docking-locking-separation mechanism; 2) detection-free passive mechanical locking; 3) magnetic repulsion-assisted separation; and 4) compact and lightweight mechanism.
These features ensure reliable docking in disturbed aerial environments.
\subsection{Self-locking Arrangement Optimization}
\begin{figure*}[!t]
\centering
    \includegraphics[width = 17 cm]{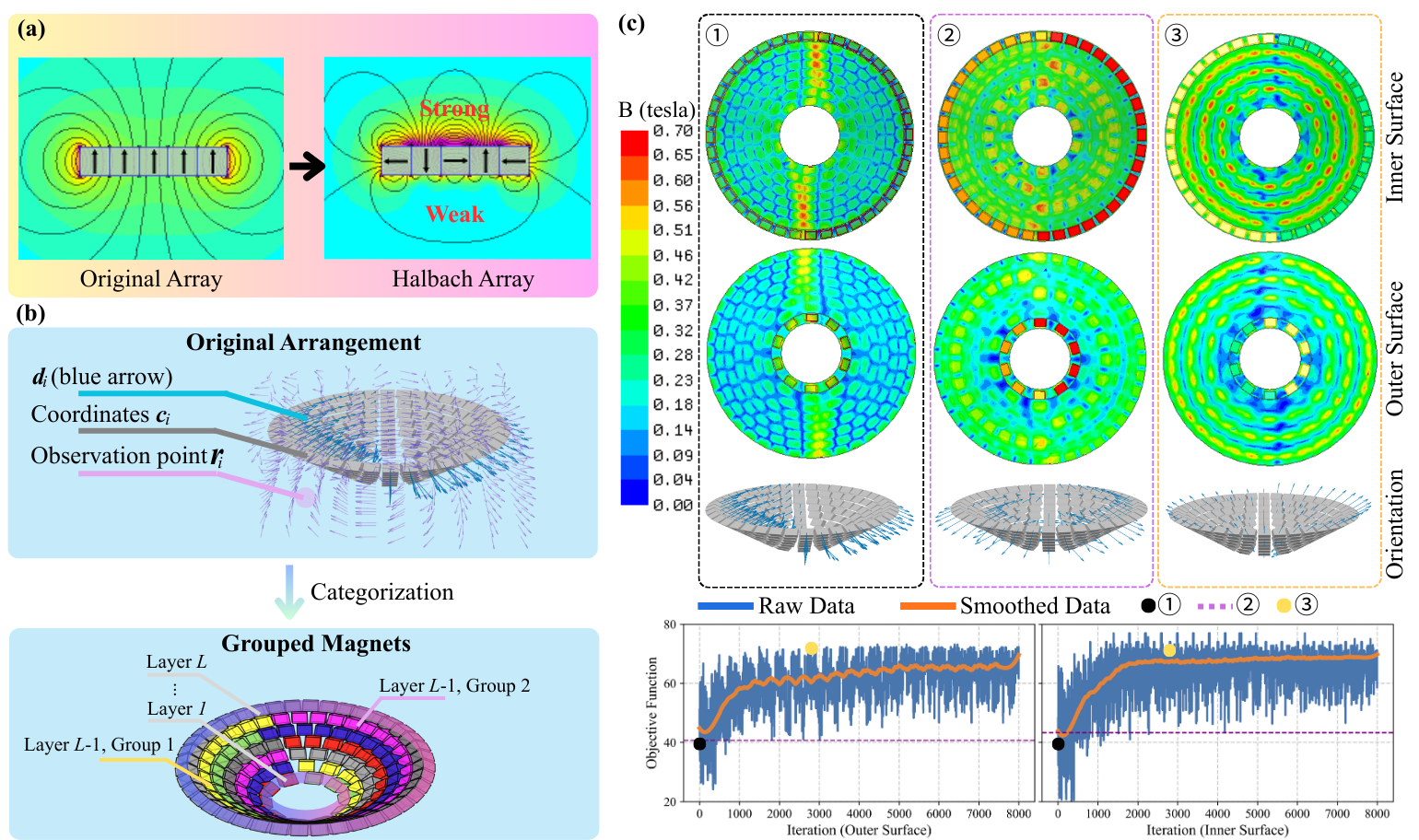}
    \vspace{-3mm}
\caption{Magnet arrangement optimization. (a) Original and Halbach array for a row of magnets. (b) The magnets are divided into $L$ layers, with each layer consisting of two groups. (c) $\textcircled{\scriptsize 1}$–$\textcircled{\scriptsize 3}$ denote the baseline, manually designed Halbach array and optimized configuration, respectively. The optimized magnet orientations and their corresponding magnetic field distributions are shown, where red regions indicate higher field intensity. The objective function values on the outer and inner surfaces versus iteration count are also presented. Configurations $\textcircled{\scriptsize 1}$, $\textcircled{\scriptsize 2}$, and $\textcircled{\scriptsize 3}$ are highlighted in black, purple, and yellow, respectively.}
\label{fig:magnet algorithm}
\vspace{-5mm}
\end{figure*}

Halbach arrays [cf. Fig.~\ref{fig:magnet algorithm}(a)] show that, by appropriately arranging magnet array structure and polarities, the magnetic field can be directed toward the desired working direction. 
This fact brings several benefits: 
1) we can optimize the magnet arrangement to enhance the magnetic force between the docking interfaces; 
and 2) given a required magnetic force, the number/weight of magnets can be optimized.
According to this fact, we formulate such magnet array arrangement optimization problem as follows.

Assume that all magnets are identical and cube-shaped. 
Denote the coordinate of the $i$-th magnet by $\boldsymbol{c}_i = [ c_{ix} \ c_{iy} \ c_{iz} ]^{\top}$, which is restricted to the docking surface.
Since the magnets are embedded in recessed slots on the docking surfaces, the magnetic moment of each magnet, denoted by $\boldsymbol{d}_i = [ d_{ix} \ d_{iy} \ d_{ix} ]^{\top}$, is restricted to only six possible orientations.
The magnet array arrangement is described by 
\begin{equation}\label{eq:magnet arrangement}
 D = \{ (\bs{c}_1, \bs{d}_1), (\bs{c}_2, \bs{d}_2), \cdots, (\bs{c}_N, \bs{d}_N) \},
\end{equation}
where $N$ is the total number of magnets.
Let $R = \{ \bs{r}_{j} \}$ be the set of observation point coordinates on the contact surface between male and female docking mechanism. 
The number of samples, i.e., observation points, is $|R|$.
The vector from the $i$-th magnet to the observation point $\bs{r}_j$ is
\[ \bs{r}_{i,j} = \bs{r}_{j} - \bs{c}_i. \]
Following the dipole approximation \cite{yurkin2007discrete}, the magnetic field vector of the $i$-th magnet observed at $\bs{r}_j$ is 
\begin{equation}\label{eq:dipole}
    \bs{B}( \bs{c}_i, \bs{d}_i, \bs{r}_{j} ) = 
    \frac{\mu_0}{4\pi}
    \left[
        \frac{3\bs{r}_{i,j}(\bs{d}_i \cdot \bs{r}_{i,j})}
        { \| \bs{r}_{i,j} \|^5}
        -\frac{\bs{d}_i}{\| \bs{r}_{i,j} \|^3}
    \right],
\end{equation}
where $\mu_0$ is the vacuum permeability constant, $\| \ \|$ is the 2-norm of a vector, and $\cdot$ denotes the dot product.
Then, the magnet array arrangement optimization problem is
\begin{equation}\label{eq:magnet optimize}
    D^{*} = \arg\max_{D} \frac{1}{|R|} \sum_{\bs{r}_j \in R} \ \sum_{(\bs{c}_i, \bs{d}_i) \in D} \| \bs{B}( \bs{c}_i, \bs{d}_i, \bs{r}_{j} ) \| ,
\end{equation}
and the corresponding magnetic field strength is
\begin{equation}\label{eq:magnet_strength}
    \bs{B}^{*} = \frac{1}{|R|} \sum_{\bs{r}_j \in R} \ \sum_{(\bs{c}_i, \bs{d}_i) \in D^{*}} \| \bs{B}( \bs{c}_i, \bs{d}_i, \bs{r}_{j} ) \| .
\end{equation}
Note that we assume that the magnet arrays of the male and female mechanisms are arranged in symmetric and opposite patterns. Thus, the magnetic fields generated by the male and female magnet arrays are symmetric, and thus in the cost function, we can simply take the magnitude of $\bs{B}$ and sum it.
In addition, we restrict the search space $R$ to points near the contact surface between the male and female docking mechanisms, since $\bs{B}( \bs{c}_i, \bs{d}_i, \bs{r}_{j} )$ decays rapidly when $\|\bs{r}_{i,j}\|$ is large.
\begin{figure}[t]
\vspace{-8pt}
\begin{algorithm}[H]
\caption{Layer-wise Magnet Arrangement Optimization}
\label{alg:magnet_opt}
\footnotesize
\begin{algorithmic}[1]
\Require initial layers $L = 0$, observation points $R$
\Statex \hspace{4.4mm} and desired magnetic field strength $\bs{B}_{\rm{d}}$
\Repeat
    \State $L \gets L + 1$
    \State set the magnet arrangement $D$ in \eqref{eq:magnet arrangement} as $L$ layers
    \Statex \hspace{4.4mm} and $2L$ groups
    \State solve \eqref{eq:magnet optimize} via multi-island genetic algorithm to obtain 
    \Statex \hspace{4.4mm} arrangement $D^{*}_{L}$ and field strength $\bs{B}^{*}_L$ as in \eqref{eq:magnet_strength}
\Until{$\bs{B}^{*}_L \geq \bs{B}_{\rm{d}}$}
\Ensure Optimized arrangement $D^{*}_{L}$
\end{algorithmic}
\end{algorithm}
\vspace{-10mm}
\end{figure}
To solve this optimization problem \eqref{eq:magnet optimize}, we employ the multi-island genetic algorithm implemented in \textit{Isight} to iteratively search for the optimal magnet configuration. 
To reduce the search space and improve convergence efficiency, the magnet array is divided into $L$ layers. 
For each layer, the magnets are partitioned into two groups of equal number, and the magnetic moments of the two groups are oriented in opposite directions, as shown in Fig.\ref{fig:magnet algorithm}(b). 
After obtaining the optimal arrangement for a given number of layers, we add the number of layers and reinitialize the optimization to search for a new arrangement. This process continues until the desired magnetic field strength is reached, as detailed in Algorithm \ref{alg:magnet_opt}.

Fig.~\ref{fig:magnet algorithm} (c) shows, in the \textit{ANSYS} simulation, the magnetic field strength distributions for optimized magnet configurations when $L=7$, along with the corresponding geometric visualizations. Fig.~\ref{fig:magnet algorithm}(c)$\textcircled{\small 1}$, $\textcircled{\small 2}$, and $\textcircled{\small 3}$ denote the baseline, the manually designed Halbach array, and the optimized configuration, respectively. 
The optimized arrangement in Fig.~\ref{fig:magnet algorithm}(c) $\textcircled{\small 3}$ achieves the highest average magnetic field strength. Compared with the baseline configuration in Fig.~\ref{fig:magnet algorithm}(c)$\textcircled{\small 1}$, it exhibits an increase of $78.80\%$ in magnetic field strength, indicating that the same number of magnets can generate a stronger magnetic force.
\section{Model Abstraction for MARS}
\label{sec:model}

In this section, We abstract MARS as a virtual quadrotor, which we refer to as the \textbf{\emph{model abstraction}}. Specifically, we introduce two types of virtual quadrotors, \textbf{\emph{equal-arm}} and \textbf{\emph{unequal-arm}}, chosen based on the intended use case.

\begin{figure}[!t]
\centering
    \includegraphics[width = 8 cm]{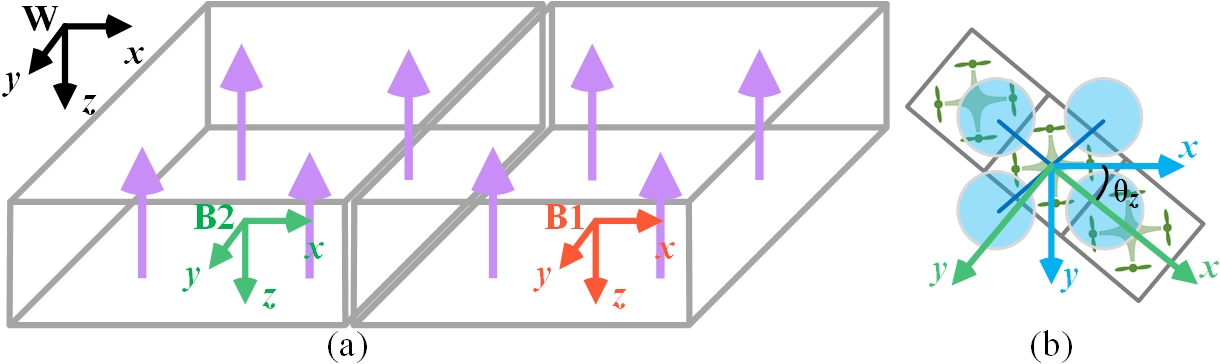}
    \vspace{-4mm}
\caption{Coordinate system representation for a MARS with two units. }
\label{fig:PX4}
\vspace{-6mm}
\end{figure}

To facilitate the subsequent formulations, we define the following notions.
The world coordinate frame or the inertial frame is $W$ as shown in Fig.~\ref{fig:PX4}(a).
The $z$-axis of the quadrotor body frame is aligned with that of the world frame.
The MARS consists of $n$ individual units.
The position of center of mass of the $i$-th unit in the world frame is denoted by $\bs{p}_i = [p_{ix} \ p_{iy} \ p_{iz}]^{\top} \in \mathbb{R}^{3}$.
The mass of the $i$-th unit is $m_i$.
Let $f_{i1}$, $f_{i2}$, $f_{i3}$, and $f_{i4}$ denote the vertical forces generated by rotors 1-4 for the $i$-th unit, 
and let $\bs{p}_{i1}$, $\bs{p}_{i2}$, $\bs{p}_{i3}$, and $\bs{p}_{i4}$ denote the positions of each rotor in the world frame. 
In particular, $\bs{p}_{ij} = [p_{ijx} \ p_{ijy} \ p_{ijz}]^{\top} \in \mathbb{R}^{3}$ for $i = 1,..,n$ and $j = 1,...,4$.

\subsection{Equal-arm Virtual Quadrotor}\label{sec:STDviture}

To determine the virtual quadrotor's structure and the control effectiveness matrix, three groups of parameters need to be specified: centroid position, orientation with respect to the physical quadrotors, and positions of the virtual rotors.


The total mass of the virtual quadrotor is
\begin{equation*}
    m_{\cV} = \sum_{i=1}^{n} m_i.
\end{equation*}
The centroid position of the virtual quadrotor in the world frame is
\begin{equation}\label{eq:centroid_position}
    \bs{p}_{\cV} = \frac{1}{nm_{\cV}} \sum_{i=1}^{n} m_i \bs{p}_i .
\end{equation}
In particular, $\bs{p}_{\cV} = [p_{\cV x} \ p_{\cV y} \ p_{\cV z}]^{\top}$.
If all units have the same mass, then $\bs{p}_{\cV} = \frac{1}{n} \sum_{i=1}^{n} \bs{p}_i$.
Note that $\bs{p}_{\cV}$ is also the centroid position of the physical MARS.

The torque $\bs{\tau}_{ij}$ generated by each rotor with respect to the centroid position of the virtual quadrotor is
\begin{equation}\label{eq:tau_ij}
    \bs{\tau}_{ij} =
    \begin{bmatrix}
        \tau_{ijx} \\ \tau_{ijy} \\ \tau_{ijz}
    \end{bmatrix}
    = 
    \begin{bmatrix}
        -p_{ijy} + p_{\cV y} \\
        p_{ijx} - p_{\cV x} \\
        (-1)^{\lfloor (j-1)/2 \rfloor + 1} c_{iz}
    \end{bmatrix}
    f_{ij},
\end{equation}
where $c_{iz}$ is a constant for $i$-th unit. $\lfloor \cdot \rfloor$ is floor operator (rounding down).
Next, we rotate the MARS about the $z$-axis at point $\bs{p}_{\cV}$ by an angle $\theta_z$, 
and the resulting torque $\bs{\tau}_{ij}(\theta)$ by each rotor is
\begin{equation}\label{eq:torque_theta_z}
    \begin{bmatrix}
        \tau_{ijx}(\theta_z) \\ \tau_{ijy}(\theta_z) \\ \tau_{ijz}
    \end{bmatrix}
    \! = \!
    \begin{bmatrix}
        \begin{bmatrix}
            \sin\theta_z & \cos\theta_z  \\
            \cos\theta_z & -\sin\theta_z
        \end{bmatrix}
        \begin{bmatrix}
            -p_{ijy} + p_{\cV y} \\
            p_{ijx} - p_{\cV x}
        \end{bmatrix}
        \\[8pt]
         (-1)^{\lfloor (j-1)/2 \rfloor + 1} c_{iz}
    \end{bmatrix}
    f_{ij} .
\end{equation}
We aim to find the optimal $\theta_z$ that maximizes the sum of the absolute values of the rotor torques, given that all rotors generate the same force. 
Without loss of generality, set $f_{ij} = 1$ for $i = 1,..,n$ and $j = 1,...,4$. 
Then we formulate the next optimization problem
\begin{equation}\label{eq:op_orientation}
    \theta_z^{*} = \arg\max_{\theta_z} \sum_{i,j} c_x | \tau_{ijx}(\theta_z) | + c_y | \tau_{ijy}(\theta_z) | 
\end{equation}
that returns the optimal orientation of the virtual quadrotor with respect to the physical MARS. Here, user-specified coefficients $c_x$ and $c_y$ balance the torque between the $x$- and $y$-axes, enabling torque compensation in asymmetric configurations and improving trajectory tracking when control authority differs across axes.

Once the optimal orientation $\theta_z^{*}$ is obtained, the final step is to determine the positions of the virtual rotors.
This step is achieved by making the virtual rotors generate the same torque as the physical MARS.
To this end, we define the following index sets
\begin{align*}
    \mathcal{I}_{x+}^{\theta_z} \! := \! \{ (i,j) : \tau_{ijx}(\theta_z) > 0 \}, \
    \mathcal{I}_{x-}^{\theta_z} \! := \! \{ (i,j) : \tau_{ijx}(\theta_z) < 0 \}, \\
    \mathcal{I}_{y+}^{\theta_z} \! := \! \{ (i,j) : \tau_{ijy}(\theta_z) > 0 \}, \
    \mathcal{I}_{y-}^{\theta_z} \! := \! \{ (i,j) : \tau_{ijy}(\theta_z) < 0 \}. 
\end{align*}
The virtual rotors with their positions $\bs{p}_{\cV j} = [p_{\cV jx} \ p_{\cV jy} \ p_{\cV z}]$ for $j=1,...,4$ should satisfy, with $f_{\rm{sum}} = \sum_{i,j} f_{ij}$,
\begin{subequations}\label{eq:rotors_position}
\begin{align}
    \sum_{ (i,j) \in \mathcal{I}_{x+}^{\theta_z^{*}} } \hspace{-3mm} \tau_{ijx}(\theta_z^{*}) & = (-p_{\cV 2y} + p_{\cV y} - p_{\cV 3y} + p_{\cV y}) \tfrac{f_{\rm{sum}}}{4} , \\
    \sum_{ (i,j) \in \mathcal{I}_{x-}^{\theta_z^{*}} } \hspace{-3mm} \tau_{ijx}(\theta_z^{*}) & = (-p_{\cV 1y} + p_{\cV y} - p_{\cV 4y} + p_{\cV y}) \tfrac{f_{\rm{sum}}}{4} , \\
    \sum_{ (i,j) \in \mathcal{I}_{y+}^{\theta_z^{*}} } \hspace{-3mm} \tau_{ijy}(\theta_z^{*}) & = (p_{\cV 1x}- p_{\cV x} + p_{\cV 3x} - p_{\cV x}) \tfrac{f_{\rm{sum}}}{4} , \\
    \sum_{ (i,j) \in \mathcal{I}_{y-}^{\theta_z^{*}} } \hspace{-3mm} \tau_{ijy}(\theta_z^{*}) & = (p_{\cV 2x} - p_{\cV x} + p_{\cV 4x} - p_{\cV x}) \tfrac{f_{\rm{sum}}}{4} .
\end{align}
\end{subequations}
The solution $(p_{\cV jx}, p_{\cV jy})$ to the above equations is the positions of the virtual rotors.
For simplicity, we set $f_{ij} = 1$ when solving \eqref{eq:rotors_position}.

Let
\begin{equation*}
    \tau_{z} = \sum_{i,j} | \tau_{ijz} | = \sum_{i,j} | (-1)^{\lfloor (j-1)/2 \rfloor + 1} c_{iz} | f_{ij},
\end{equation*}
and take
\begin{equation}\label{eq:std_kvz}
    c_{\cV z} = \tau_{z} / f_{\rm{sum}},
\end{equation}
where $f_{ij} = 1$ for simplicity.
Then we obtain the control effectiveness matrix $\bs{G}_{\cV}$ for the virtual quadrotor as:
\begin{equation}\label{eq:ctrl_eff_matrix_std}
    \bs{G}_{\cV} \! = \!
    \left[\begin{smallmatrix}
        1 & 1 & 1 & 1 \\
        -p_{\cV 1y} + p_{\cV y} & -p_{\cV 2y} + p_{\cV y} & -p_{\cV 3y} + p_{\cV y} & -p_{\cV 4y} + p_{\cV y} \\
        p_{\cV 1x} - p_{\cV x} & p_{\cV 2x} - p_{\cV x} & p_{\cV 3x} - p_{\cV x} & p_{\cV 4x} - p_{\cV x} \\
        - c_{\cV z} & -c_{\cV z} & c_{\cV z} & c_{\cV z} 
    \end{smallmatrix}\right],
\end{equation}
where $\bs{p}_{\cV}$ is from \eqref{eq:centroid_position}, $\bs{p}_{\cV 1}$,...,$\bs{p}_{\cV 4}$ are from \eqref{eq:rotors_position}, and $c_{\cV z}$ is from \eqref{eq:std_kvz}.

In summary, the procedure to determine the virtual quadrotor's structure is stated in Algorithm \ref{alg:StandardVirtual}.
\begin{figure}[t]
\vspace{-8pt}
\begin{algorithm}[H]
\caption{Equal-arm Virtual Quadrotor Structure Design}
\label{alg:StandardVirtual}
\footnotesize
\begin{algorithmic}[1]   
    \Require unit mass $m_i$, unit position $\bs{p}_i$, coefficient $c_x$ and $c_y$,
    \Statex \hspace{3.6mm} rotor position $\bs{p}_{ij}$, rotor force $f_{ij}=1$
    \State compute virtual quadrotor centroid position $\bs{p}_{\cV}$ by \eqref{eq:centroid_position}
    \State compute optimal orientation $\theta_z^{*}$ by \eqref{eq:op_orientation}
    \State compute virtual rotors positions $\bs{p}_{\cV 1}$,...,$\bs{p}_{\cV· 4}$ by \eqref{eq:rotors_position}
    \State compute torque coefficient $c_{\cV z}$ by \eqref{eq:std_kvz}
    \Ensure virtual quadrotor's structure $\bs{p}_{\cV}$, $\theta_z^{*}$, and $\bs{p}_{\cV 1}$,...,$\bs{p}_{\cV 4}$,
    \Statex \hspace{5.7mm} and control effectiveness matrix $\bs{G}_{\cV}$
\end{algorithmic}
\end{algorithm}
\vspace{-10mm}
\end{figure}
\begin{remark}
To handle the absolute value term in the objective function \eqref{eq:op_orientation}, we proceed as follows.
Define the following notations:
\begin{subequations}\label{eq:f_pmxy}
\begin{align}
    \bs{f} & \! = \! 
    \left[\begin{smallmatrix}
        f_{11} \cdots f_{14} & f_{21} \cdots f_{24} \cdots f_{n1} \cdots f_{n4}
    \end{smallmatrix}\right]
    \in \mathbb{R}^{1 \times 4n}, 
    \\
    \bs{p}_{\cM x} & \! = \!  
    \left[\begin{smallmatrix}
        p_{11x} \cdots p_{14x} & p_{21x} \cdots p_{24x} \cdots p_{n1x} \cdots p_{n4x}
    \end{smallmatrix}\right]
    \! - \! p_{\cV x} \mathbf{1}_{1 \times 4n} , 
    \\
    \bs{p}_{\cM y} & \! = \!  
    \left[\begin{smallmatrix}
        p_{11y} \cdots p_{14y} & p_{21y} \cdots p_{24y} \cdots p_{n1y} \cdots p_{n4y}
    \end{smallmatrix}\right]
    \! - \! p_{\cV y} \mathbf{1}_{1 \times 4n},
\end{align}
\end{subequations}
where $\mathbf{1}_{a \times b}$ denotes an $a \times b$ matrix whose entries are all ones; $p_{\cV x}$ and $p_{\cV y}$ are the $x$- and $y$-coordinates of the center of mass as in \eqref{eq:centroid_position}.
The above notations satisfy
\begin{align}
    & \begin{bmatrix}
        {\tau}_{\bs x +}(\theta_z) \\ 
        {\tau}_{\bs y +}(\theta_z) \\ 
        {\tau}_{\bs x -}(\theta_z) \\ 
        {\tau}_{\bs y -}(\theta_z)
    \end{bmatrix}
    =  
    \begin{bmatrix}
        \sum_{ (i,j) \in \mathcal{I}_{x+}^{\theta_z} } \tau_{ijx}(\theta_z) \\ 
        \sum_{ (i,j) \in \mathcal{I}_{y+}^{\theta_z} } \tau_{ijx}(\theta_z) \\ 
        \sum_{ (i,j) \in \mathcal{I}_{x-}^{\theta_z} } \tau_{ijx}(\theta_z) \\
        \sum_{ (i,j) \in \mathcal{I}_{y-}^{\theta_z} } \tau_{ijx}(\theta_z)
    \end{bmatrix} 
    \label{eq:tau_plus_minus}
    \\
    & = 
    \begin{bmatrix}
        \max\left(    
            \begin{bmatrix}
                \sin\theta_z & \cos\theta_z\\
                \cos\theta_z & -\sin\theta_z
            \end{bmatrix}
            \begin{bmatrix}
                -\bs{p}_{\cM y} \\ \bs{p}_{\cM x}
            \end{bmatrix} ,
            \mathbf{0}_{2 \times 4n}
        \right)
        \bs{f}^{\top}
        \\[8pt]
        \min\left(    
            \begin{bmatrix}
                \sin\theta_z & \cos\theta_z\\
                \cos\theta_z & -\sin\theta_z
            \end{bmatrix}
            \begin{bmatrix}
                -\bs{p}_{\cM y} \\ \bs{p}_{\cM x}
            \end{bmatrix} ,
            \mathbf{0}_{2 \times 4n}
        \right)
        \bs{f}^{\top}
    \end{bmatrix} 
    \notag,
\end{align}
where $\mathbf{0}_{a \times b}$ denotes an $a \times b$ matrix whose entries are all zeros; $\min(\cdot)$ and $\max(\cdot)$ denote element-wise operations that select the minimum and maximum values for each corresponding element of the input vectors.
Thanks to the expression \eqref{eq:tau_plus_minus}, the optimization problem \eqref{eq:op_orientation} rewrites as
\begin{align}\label{eq:op_orientation_re}
    \theta_z^{*} = \arg\max_{\theta_z}  
    [ c_x {\tau}_{\bs x +}(\theta_z) 
    & - c_x {\tau}_{\bs x -}(\theta_z)
    \notag
    \\[-4pt]
    & + c_y {\tau}_{\bs y +}(\theta_z) 
    - c_y {\tau}_{\bs y -}(\theta_z)].
\end{align}
This problem is solved numerically using CasADi~\cite{andersson2019casadi}. 
\hfill $\triangleleft$
\end{remark}
\subsection{Unequal-arm Virtual Quadrotor}
In the equal-arm virtual quadrotor model, the virtual propellers/rotors are arranged across four distinct quadrants, forming a typical $\times$-shaped configuration. 
To enhance flexibility, we introduce a unequal-arm virtual quadrotor model, which allows the rotors to be placed at arbitrary locations. Similar to the equal-arm model abstraction, we formulate an optimization problem to maximize the torque of the virtual quadrotor in this unequal-arm case. 
To this end, we first denote the minimum and maximum values of each rotor force as $\underline{f}_{ij}$ and $\bar{f}_{ij}$, respectively. Then
\begin{equation*}
    f_{ij} \in \left[ \underline{f}_{ij} \, , \ \bar{f}_{ij} \right],
\end{equation*}
and 
\begin{equation*}
    f_{\rm{sum}} \in \left[ \underline{f}_{\rm{sum}} \,, \ \bar{f}_{\rm{sum}} \right], \ 
    \underline{f}_{\rm{sum}} = \sum_{ij} \underline{f}_{ij}, \ 
    \bar{f}_{\rm{sum}} = \sum_{ij} \bar{f}_{ij} .
\end{equation*}
As $i = 1,...,n$ and $j = 1,...,4$, there are $4n$ rotors, and each rotor can take either its minimum or maximum force, resulting in $2^{4n}$ possible combinations in total.
Arrange these combinations into a matrix
\begin{equation*}
    \bs{V}_{\cM} = 
    \begin{bmatrix}
        \underline{f}_{11} & \bar{f}_{11} & \underline{f}_{11} & \bar{f}_{11} & \cdots & \bar{f}_{11} \\[4pt]
        \underline{f}_{12} & \underline{f}_{12} & \bar{f}_{12} & \bar{f}_{12} & \cdots & \bar{f}_{12} \\
        \vdots & \vdots & \vdots & \vdots & \ddots & \vdots \\
        \underline{f}_{n3} & \underline{f}_{n3} & \underline{f}_{n3} & \underline{f}_{n3} & \cdots & \bar{f}_{n3} \\[4pt]
        \underline{f}_{n4} & \underline{f}_{n4} & \underline{f}_{n4} & \underline{f}_{n4} & \cdots & \bar{f}_{n4}
    \end{bmatrix}
    \in \mathbb{R}^{4n \times 2^{4n}}.
\end{equation*}
In a similar manner, we can define 
\begin{equation*}
    \bs{V}_{\cV} = 
    \frac{1}{4}
    \begin{bmatrix}
        \underline{f}_{\rm{sum}} & \bar{f}_{\rm{sum}} & \underline{f}_{\rm{sum}} & \bar{f}_{\rm{sum}} \cdots \bar{f}_{\rm{sum}} \\[4pt]
        \underline{f}_{\rm{sum}} & \underline{f}_{\rm{sum}} & \bar{f}_{\rm{sum}} & \bar{f}_{\rm{sum}} \cdots \bar{f}_{\rm{sum}} \\[4pt]
        \underline{f}_{\rm{sum}} & \underline{f}_{\rm{sum}} & \underline{f}_{\rm{sum}} & \underline{f}_{\rm{sum}} \cdots \bar{f}_{\rm{sum}} \\[4pt]
        \underline{f}_{\rm{sum}} & \underline{f}_{\rm{sum}} & \underline{f}_{\rm{sum}} & \underline{f}_{\rm{sum}} \cdots \bar{f}_{\rm{sum}}
    \end{bmatrix}
    \in \mathbb{R}^{4 \times 2^{4}} ,
\end{equation*}
which is the combinations of the minimum or maximum force by the virtual rotors.
The control effectiveness matrix for the MARS is 
\begin{equation}\label{eq:matrixGM}
    \bs{G}_{\cM} \! = \! 
    \begin{bmatrix}
        \mathbf{1}_{1 \times 4n} \\
        -\bs{p}_{\cM y} \\
        \bs{p}_{\cM x} \\
        -c_{1z} \ ... \ (-1)^{\lfloor (j-1)/2 \rfloor + 1} c_{iz} \ ... \  c_{nz}
    \end{bmatrix}
    \! \in \! \mathbb{R}^{4 \times 4n},
\end{equation}
where $\bs{p}_{\cM x}$ and $\bs{p}_{\cM y}$ are defined in \eqref{eq:f_pmxy}; $(-1)^{\lfloor (j-1)/2 \rfloor + 1} c_{iz}$ is the same as in \eqref{eq:tau_ij} for $i = 1,..,n$ and $j = 1,...,4$.
The control effectiveness matrix for the virtual quadrotor is
\begin{equation}\label{eq:matrixGV}
    \bs{G}_{\cV} =
    \left[\begin{smallmatrix}
        1 & 1 & 1 & 1 \\
        -p_{\cV 1y} + p_{\cV y} & -p_{\cV 2y} + p_{\cV y} & -p_{\cV 3y} + p_{\cV y} & -p_{\cV 4y} + p_{\cV y} \\
        p_{\cV 1x} - p_{\cV x} &  p_{\cV 2x} - p_{\cV x} & p_{\cV 3x} - p_{\cV x} & p_{\cV 4x} - p_{\cV x} \\
        - c_{\cV z} & -c_{\cV z} & c_{\cV z} & c_{\cV z} 
    \end{smallmatrix}\right].
\end{equation}
Note that $[p_{\cV x} \ p_{\cV y} \ p_{\cV z}]^{\top} = \bs{p}_{\cV}$ is obtained from \eqref{eq:centroid_position}; 
virtual rotors positions $\bs{p}_{\cV j} = [p_{\cV jx} \ p_{\cV jy} \ p_{\cV z}]$ for $j=1,...,4$ and parameter $c_{\cV z}$ in $\bs{G}_{\cV}$ is to be determined later.

The MARS's total thrust and torque vertex matrix is $\bs{G}_{\cM} \bs{V}_{\cM}$; 
the virtual quadrotor's total thrust and torque vertex matrix is $\bs{G}_{\cV} \bs{V}_{\cV}$;
$\bs{G}_{\cV} \bs{V}_{\cV}$ should be the convex combination of $\bs{G}_{\cM} \bs{V}_{\cM}$;
meanwhile, the virtual quadrotor should generate the maximum torque. 
Thus, we formulate the cost function as
\begin{align*}
    c(\bs{p}_{\cV 1},...,\bs{p}_{\cV 4}) 
    = \tfrac{\bar{f}_{\rm{sum}}}{4} \Big[
    & \left( -p_{\cV 2y} + p_{\cV y} \right) + \left( -p_{\cV 3y} + p_{\cV y} \right)
    \\
    -& \left( -p_{\cV 1y} + p_{\cV y} \right) - \left(-p_{\cV 4y} + p_{\cV y} \right)
    \\
    + & \left( p_{\cV 1x} - p_{\cV x} \right) + \left( p_{\cV 3x} - p_{\cV x} \right)
    \\
    - & \left( p_{\cV 2x} - p_{\cV x} \right) - \left( p_{\cV 4x} - p_{\cV x} \right) \Big],
\end{align*}
and formulate the optimization problem as
\begin{subequations}\label{eq:flex_op_plbm}
\begin{align}
    & \max_{\bs{p}_{\cV 1},...,\bs{p}_{\cV 4},c_{\cV z},\bs{\alpha}} c(\bs{p}_{\cV 1},...,\bs{p}_{\cV 4}) 
    \label{eq:flex_op_plbm_obj}
    \\
    & \rm{s.t.} \ \bs{G}_{\cV} \bs{V}_{\cV} = \bs{G}_{\cM} \bs{V}_{\cM} \ \bs{\alpha}, \ \
    \bs{\alpha} = [\alpha_{\iota\varsigma}]_{2^{4n} \times 2^{4}}    \label{eq:cvx_combi}
    \\
    & \hspace{7mm} \alpha_{\iota\varsigma} \geq 0 \text{ for } \iota = 1,...,2^{4n}, \ \varsigma = 1,...,2^{4}
    \\
    & \hspace{7mm} \sum_{\iota = 1}^{2^{4n}} \alpha_{\iota\varsigma} = 1 \text{ for } \varsigma = 1,...,2^{4} .
\end{align}
\end{subequations}
In summary, the procedure to determine the virtual quadrotor's structure is stated in Algorithm \ref{alg:FlexibleVirtual}.

\begin{figure}[t]
\vspace{-8pt}
\begin{algorithm}[H]
\caption{Unequal-arm Virtual Quadrotor Structure Design}
\footnotesize
\label{alg:FlexibleVirtual}
\begin{algorithmic}[1]   
    \Require unit mass $m_i$, unit position $\bs{p}_i$,
    \Statex \hspace{3.7mm} rotor position $\bs{p}_{ij}$, rotor force vertex $\underline{f}_{ij}$ and $\bar{f}_{ij}$
    \State compute virtual quadrotor centroid position $\bs{p}_{\cV}$ by \eqref{eq:centroid_position}
    \State compute virtual rotors positions $\bs{p}_{\cV 1}$,...,$\bs{p}_{\cV 4}$ and torque coefficient $c_{\cV z}$ by \eqref{eq:flex_op_plbm}
    \Ensure virtual quadrotor's structure $\bs{p}_{\cV}$ and $\bs{p}_{\cV 1}$,...,$\bs{p}_{\cV 4}$, and 
    \Statex \hspace{5.7mm} optimal control effectiveness matrix $\bs{G}_{\cV}$
\end{algorithmic}
\end{algorithm}
\vspace{-9mm}
\end{figure}

\begin{remark}
If, for example, there are 4 units in the MARS, the number of vertices is $2^{4 \times 4} = 65536$,
leading to the decision variable $\bs{\alpha}$ of size $16 \times 65536$ in \eqref{eq:flex_op_plbm}.
The computational complexity grows with the number of decision variables.
To reduce the number of variables  and accelerate the computation of this optimization problem, we introduce the following simplifications.
Let, for $i=1,...,n$,
\begin{equation*}
    \underline{f}_{i} = \underline{f}_{i1} + \underline{f}_{i2} + \underline{f}_{i3} + \underline{f}_{i4}, \
    \bar{f}_{i} = \bar{f}_{i1} + \bar{f}_{i2} + \bar{f}_{i3} + \bar{f}_{i4},
\end{equation*}
and 
\begin{equation*}
    \bs{V}^{\prime}_{\cM} 
    =
    \begin{bmatrix}
        \underline{f}_{1} & \bar{f}_{1} & \underline{f}_{1} & \bar{f}_{1} & \cdots & \bar{f}_{1} \\[4pt]
        \underline{f}_{2} & \underline{f}_{2} & \bar{f}_{2} & \bar{f}_{2} & \cdots & \bar{f}_{2} \\
        \vdots & \vdots & \vdots & \vdots & \ddots & \vdots \\
        \underline{f}_{n} & \underline{f}_{n} & \underline{f}_{n} & \underline{f}_{n} & \cdots & \bar{f}_{n}
    \end{bmatrix}
    \in \mathbb{R}^{n \times 2^n} .
\end{equation*}
The control effectiveness matrix of the MARS is
\begin{equation*}
    \bs{G}^{\prime}_{\mathcal{M}} 
    = 
    \left[\begin{smallmatrix}
        1 & 1 & \cdots & 1 \\
        -p_{1y} + p_{\cV y} & -p_{2y} + p_{\cV y} & \cdots & -p_{ny} + p_{\cV y} \\
        p_{1x} - p_{\cV x} & p_{2x} - p_{\cV x} & \cdots & p_{nx} - p_{\cV x} \\
        0 & 0 & \cdots & 0
    \end{smallmatrix}\right] \in \mathbb{R}^{4 \times n},
\end{equation*}
where $(p_{ix},p_{iy})$ is the $(x,y)$ position of $i$-th unit.
Then the optimization problem \eqref{eq:flex_op_plbm} reduces to 
\begin{subequations}\label{eq:flex_op_plbm_re}
\begin{align}
    & \max_{\bs{p}_{\cV 1},...,\bs{p}_{\cV 4},c_{\cV z},\bs{\alpha}} c(\bs{p}_{\cV 1},...,\bs{p}_{\cV 4}) 
    \\
    & \rm{s.t.} \ \bs{G}_{\cV} \bs{V}_{\cV} = \bs{G}_{\cM}^{\prime} \bs{V}_{\cM}^{\prime} \ \bs{\alpha}^{\prime}, \ \
    \bs{\alpha}^{\prime} = [\alpha_{\iota\varsigma}^{\prime}]_{2^{n} \times 2^{4}}    
    \\
    & \hspace{7mm} \alpha_{\iota\varsigma}^{\prime} \geq 0 \text{ for } \iota = 1,...,2^{n}, \ \varsigma = 1,...,2^{4}
    \\
    & \hspace{7mm} \sum_{\iota = 1}^{2^{n}} \alpha_{\iota\varsigma}^{\prime} = 1 \text{ for } \varsigma = 1,...,2^{4} .
\end{align}
\end{subequations}
Moreover, $c_{\cV z}$ is computed via \eqref{eq:std_kvz} by taking $f_{ij} = \bar{f}_{ij}$ in this simplification. 
To sum up, solving \eqref{eq:centroid_position}, \eqref{eq:flex_op_plbm_re} and \eqref{eq:std_kvz} yields the virtual quadrotor's structure $\bs{p}_{\cV}$ and $\bs{p}_{\cV 1}$,...,$\bs{p}_{\cV 4}$, and control effectiveness matrix $\bs{G}_{\cV}$.
\hfill $\triangleleft$
\end{remark}
\subsection{Discussion}
Equal-arm and unequal-arm virtual quadrotor model abstractions have their own benefits and limitations.
For the equal-arm model, net torque generated by the virtual quadrotor matches that of the original MARS configuration. 
In other words, the constrain \eqref{eq:rotors_position} enforces that there is no approximation error between the physical MARS and the virtual quadrotor.
However, the force required by the virtual quadrotor are not always realizable by the physical MARS, namely,
given the control effectiveness matrix $\bs{G}_{\cV}$ solved from Algorithm \ref{alg:StandardVirtual} and \eqref{eq:ctrl_eff_matrix_std}, $\bs{G}_{\cV} \bs{V}_{\cV}$ is not necessarily the convex combination of $\bs{G}_{\cM} \bs{V}_{\cM}$.
For the unequal-arm model, the constrain \eqref{eq:cvx_combi} guarantees that the force required by the virtual quadrotor can be recovered from the physical MARS.
However, there exists approximation error, and such error can be quantified as
\begin{align}\label{eq:approxi_error}
    e_{\rm{approx},1} & = 
    \tfrac{(-p_{\cV 2y} + p_{\cV y} - p_{\cV 3y} + p_{\cV y}) \tfrac{f_{\rm{sum}}}{4} - \sum_{ (i,j) \in \mathcal{I}_{x+}^{0} } \tau_{ijx}(0)}{\sum_{ (i,j) \in \mathcal{I}_{x+}^{0} } \tau_{ijx}(0)} ,
    \notag
    \\
    e_{\rm{approx},2} & =
    \tfrac{(-p_{\cV 1y} + p_{\cV y} - p_{\cV 4y} + p_{\cV y}) \tfrac{f_{\rm{sum}}}{4} - \sum_{ (i,j) \in \mathcal{I}_{x-}^{0} } \tau_{ijx}(0)}{\sum_{ (i,j) \in \mathcal{I}_{x-}^{0} } \tau_{ijx}(0)} ,
    \notag
    \\
    e_{\rm{approx},3} & = 
    \tfrac{(p_{\cV 1x}- p_{\cV x} + p_{\cV 3x} - p_{\cV x}) \tfrac{f_{\rm{sum}}}{4} - \sum_{ (i,j) \in \mathcal{I}_{y+}^{0} } \tau_{ijy}(0)}{\sum_{ (i,j) \in \mathcal{I}_{y+}^{0} } \tau_{ijy}(0)} ,
    \notag
    \\
    e_{\rm{approx},4} & = 
    \tfrac{(p_{\cV 2x} - p_{\cV x} + p_{\cV 4x} - p_{\cV x}) \tfrac{f_{\rm{sum}}}{4} - \sum_{ (i,j) \in \mathcal{I}_{y-}^{0} } \tau_{ijy}(0)}{\sum_{ (i,j) \in \mathcal{I}_{y-}^{0} } \tau_{ijy}(0)} ,
    \notag
    \\
    e_{\rm{approx}} & =  \frac{1}{4} \big( |e_{\rm{approx},1}| + ... + |e_{\rm{approx},4}| \big) ,
\end{align}
where $\bs{p}_{\cV}$ and $\bs{p}_{\cV 1}$,...,$\bs{p}_{\cV 4}$ are returned from Algorithm \ref{alg:FlexibleVirtual}, $f_{ij}$ is set to $\bar{f}_{ij}$, and $f_{\rm{sum}}$ is set to $\bar{f}_{\rm{sum}}$.

\section{Abstracted Predictive Control and Allocation}
\label{sec:controller}
\begin{figure*}[!t]
\centering
    \includegraphics[width = 17 cm]{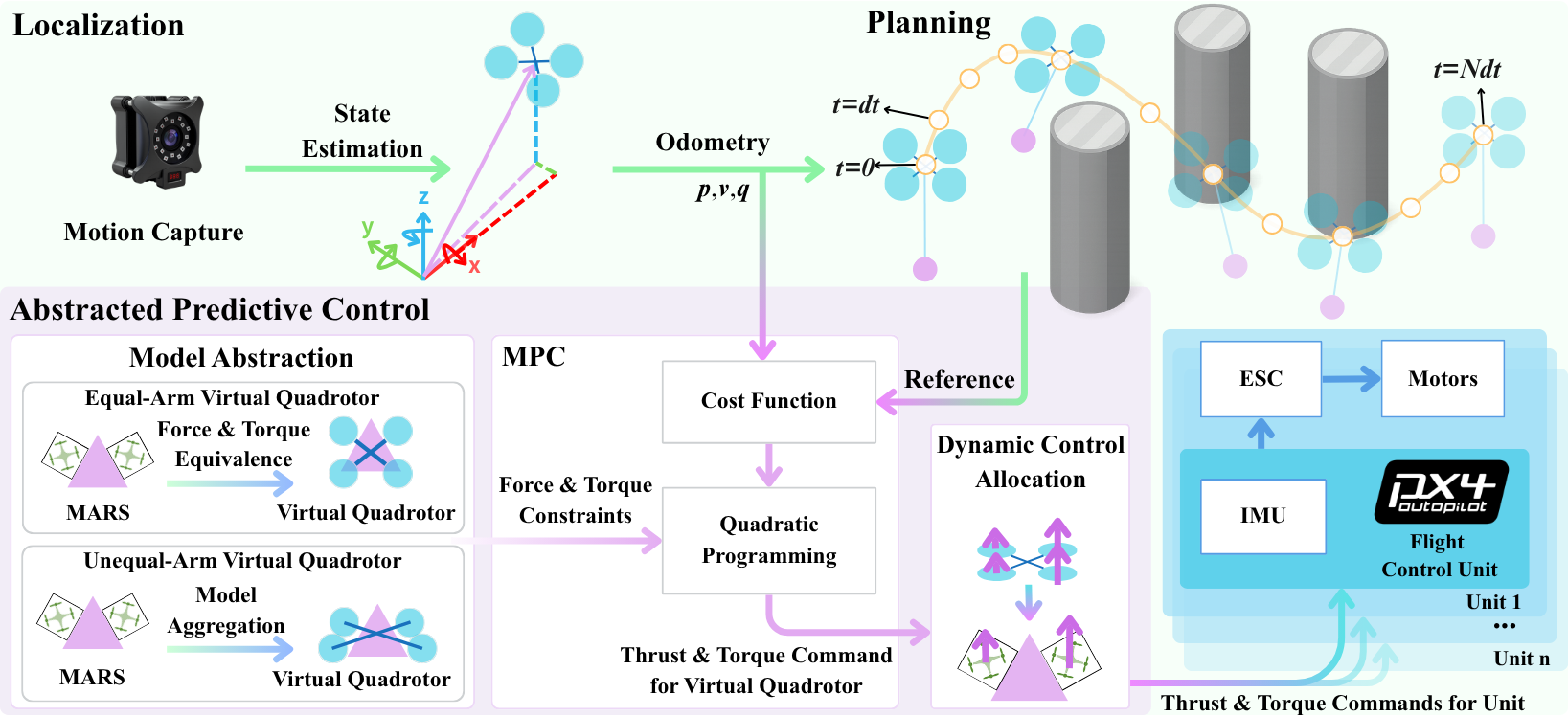}
    \vspace{-3mm}
\caption{Model abstraction and APC framework for MARS. Localization provides odometry for high-level planning. Model abstraction converts a given MARS configuration into a virtual quadrotor, which can be either an equal-arm model or an unequal-arm model, and can be extended to configurations carrying external loads. The abstracted dynamic model and control constraints are incorporated into an APC, which computes real-time control inputs for the virtual quadrotor. An optimization-based dynamic control allocation then maps the virtual control commands to thrust and torque commands for all drone units.}
\label{fig:APC}
\vspace{-6mm}
\end{figure*}
We present a two-stage abstracted predictive–allocation controller, as illustrated in Fig.~\ref{fig:APC}. In the first stage, virtual control inputs are computed by minimizing deviations from desired state trajectories subject to the virtual quadrotor dynamics and bounded force and torque constraints. In the second stage, these virtual control inputs are mapped to individual drone units in MARS through balanced dynamic control allocation, enabling scalable operation under arbitrary MARS configurations and avoiding the discontinuous control commands in prior work~\cite{Modquad,ModQuad-Vi,Modquad-gripper}. We introduce the method through its model, controller, and control allocation components.

\subsection{Virtual Quadrotor Modeling}
We use the parameters obtained from the model abstraction in the previous section. 
Prior formulations~\cite{Modquad, ModQuad-Vi, Modquad-dof, saldana2019design} rely on a simplified model valid only near hovering, which limits agility and often leads to oscillatory behavior. 
In contrast, our dynamic virtual quadrotor model captures the full system dynamics and explicitly incorporates rotor force limits, enabling stable and agile operation.

The quadrotor body frame is shown in Fig.~\ref{fig:PX4}. 
The virtual quadrotor dynamics in quaternion form are
\begin{equation}
\begin{aligned}&\dot{\boldsymbol{p}}=\boldsymbol{v},
&&\dot{\boldsymbol{q}}=\frac12\boldsymbol{q}\circ\begin{bmatrix}0\\\boldsymbol{\omega}\end{bmatrix},\\
&\dot{\boldsymbol{v}}=\boldsymbol{q}\odot\begin{bmatrix}0\\0\\\frac{F}{m_\mathcal{V}}\end{bmatrix}-\boldsymbol{g},
&&\dot{\boldsymbol{\omega}}=\boldsymbol{I}_\mathcal{V}^{-1}\left(\boldsymbol{\boldsymbol{M}}-\boldsymbol{\omega}\times(\boldsymbol{I}_\mathcal{V}\boldsymbol{\omega})\right),
\label{eq:APC_model}
\end{aligned}
\end{equation}
where $\boldsymbol{p}$ and $\boldsymbol{v}$ denote the position and velocity of the center of mass in the inertial frame; $\boldsymbol{q}=[q_w \ q_x \ q_y \ q_z]^\top$ represents orientation of the virtual quadrotor; $\boldsymbol{\omega}=[ \omega_x \ \omega_y \ \omega_z]^\top$ is the angular velocity of the virtual quadrotor in the body frame; $\boldsymbol{g}=[0 \ 0 \ g]^\top$ is the gravitational acceleration vector; $\boldsymbol{I}_\mathcal{V}$ is the inertia matrix of the entire virtual quadrotor that can be calculated using the parallel axis theorem. The operator $\circ$ denotes the quaternion multiplication, and $\odot$ denotes the rotation of a vector using a quaternion. $F$ and $\boldsymbol{\boldsymbol{M}}$ are collective thrust and torque generated by actuators. 

\subsection{Model Predictive Control}
Let $\bs{x} = [\bs{p}^{\top} \ \bs{q}^{\top} \ \bs{v}^{\top} \ \bs{\omega}^{\top}]^{\top}$ and $\bs{u}_{\cV} = [F \ \bs{M}^{\top}]^{\top}$. 
The abstracted virtual quadrotor dynamics can be expressed as
\[ \boldsymbol{x}_{k+1}=f(\boldsymbol{x}_{k},\boldsymbol{u}_{\mathcal{V},k}). \]



%

Given a reference trajectory $\boldsymbol{x}_{k:k+N-1}^{\tref}$, the APC is to compute the optimal control sequence $\boldsymbol{u}_{\mathcal{V},k:k+N-1}$ that minimizes the cost function while satisfying the system dynamics and constraints:
\begin{align}\label{eq:APC}
    \min_{\bs{u}_{\cV, k:k+N-1}} \hspace{-1mm}
    & \bs{y}_{k+N}^\top \bs{Q}_N \bs{y}_{k+N} 
    + \hspace{-3mm}
    \sum_{i=k}^{k+N-1} \hspace{-2mm}
    \left( \bs{y}_i^\top \bs{Q} \bs{y}_i + \bs{u}_{\cV,i}^\top \bs{R} \bs{u}_{\cV,i} \right)
    \notag
    \\
    \text{s.t. } \ 
    & \bs{x}_{i+1}=f(\bs{x}_i,\bs{u}_{\cV,i}) \quad i=k,k+1,\ldots,k+N-1
    \notag
    \\
    & \underline{\bs{u}}_{\cV} \leq \bs{u}_{\cV} \leq \bar{\bs{u}}_{\cV}
\end{align}
where $\bs{y}_i = \bs{x}_{i} - \bs{x}_{i}^{\tref}$; 
$k$ is the current time step; 
$N$ is the prediction horizon.
The weighting matrices \( \bs{Q} \), \( \bs{Q}_N \), and \( \bs{R} \) are defined as
\begin{subequations}
\begin{align}
    \boldsymbol{Q} & = 
    \mathrm{diag} \left( \bs{Q}_{\bs{p}}, \bs{Q}_{\bs{q}}, \bs{Q}_{\bs{v}}, \bs{Q}_{\bs{\omega}} \right),\\
    \boldsymbol{Q}_N & =
    \mathrm{diag}\left( \bs{Q}_{\bs{p},N}, \bs{Q}_{\bs{q},N}, \bs{Q}_{\bs{v},N}, \bs{Q}_{\bs{\omega},N} \right),\\
    \boldsymbol{R} & = 
    \mathrm{diag}\left( R_{F}, R_{M_x}, R_{M_y}, R_{M_z} \right),
\end{align}
\end{subequations}
where $\mathrm{diag}(\cdot)$ denotes a diagonal matrix with the specified elements on its main diagonal.
The lower and upper bounds of the virtual actuator are set to
\begin{align*}
    \underline{\bs{u}}_{\cV} & = \tfrac{1}{4} \bs{G}_{\cV} 
    \begin{bmatrix}
        \underline{f}_{\rm{sum}} & \underline{f}_{\rm{sum}} & \underline{f}_{\rm{sum}} & \underline{f}_{\rm{sum}}
    \end{bmatrix}^{\top}
    \\
    \bar{\bs{u}}_{\cV} & = \tfrac{1}{4} \bs{G}_{\cV} 
    \begin{bmatrix}
        \bar{f}_{\rm{sum}} & \bar{f}_{\rm{sum}} & \bar{f}_{\rm{sum}} & \bar{f}_{\rm{sum}}
    \end{bmatrix}^{\top} 
\end{align*}
where $\bs{G}_{\cV}$ is solved from Algorithm \ref{alg:StandardVirtual} for the equal-arm model or from Algorithm \ref{alg:FlexibleVirtual} for the unequal-arm model, and $\underline{f}_{\rm{sum}}$ ($\bar{f}_{\rm{sum}}$) is the sum of the minimum (maximum) forces of all rotors.

\subsection{Balanced Dynamic Control Allocation}

The control command $\bs{u}_{\cV,k}$ obtained from \eqref{eq:APC} is then mapped to each individual unit rotor force $f_{ij,k}$.
With a slight abuse of notation
\[ \bs{f}_{k} = [f_{11,k} \cdots f_{14,k} \ f_{21,k} \cdots f_{24,k} \cdots f_{n1,k} \cdots f_{n4,k}], \]
we formulate the following optimization problem running at each control step:
\begin{subequations}\label{eq:op_allocate}
\begin{align}
    & \min_{\bs{f}_{k}} \ {\rm{Var}}(\bs{f}_{k})
    \\
    \text{s.t. } & \bs{u}_{\cV,k} = \bs{G}_{\cM} \bs{f}_{k}^{\top} , \quad
    \underline{f}_{ij} \leq f_{ij,k} \leq \bar{f}_{ij} 
    \\
    & \text{for } i=1,...,n \text{ and } j=1,...,4
    \notag
\end{align}
\end{subequations}
where $\bs{G}_{\cM}$ is from \eqref{eq:matrixGM}, and the variance $\rm{Var}(\cdot)$ is 
\begin{equation*}
    \mathrm{Var}(\bs{f_k}) = \frac{1}{4n} \sum_{i,j} \Big( f_{ij,k} - \frac{1}{4n} \sum_{i,j} f_{ij,k} \Big)^2 .
\end{equation*}

Minimizing rotor-force variance prevents unbalanced lift distribution that would otherwise cause outer motors to idle and inner motors to overload, while yielding continuous control commands that enable scalable operation under arbitrary MARS configurations and avoid the discontinuities seen in prior work~\cite{Modquad,ModQuad-Vi,Modquad-gripper}.

\begin{figure}[!t]
    \centering
    \includegraphics[width= 0.49 \textwidth]{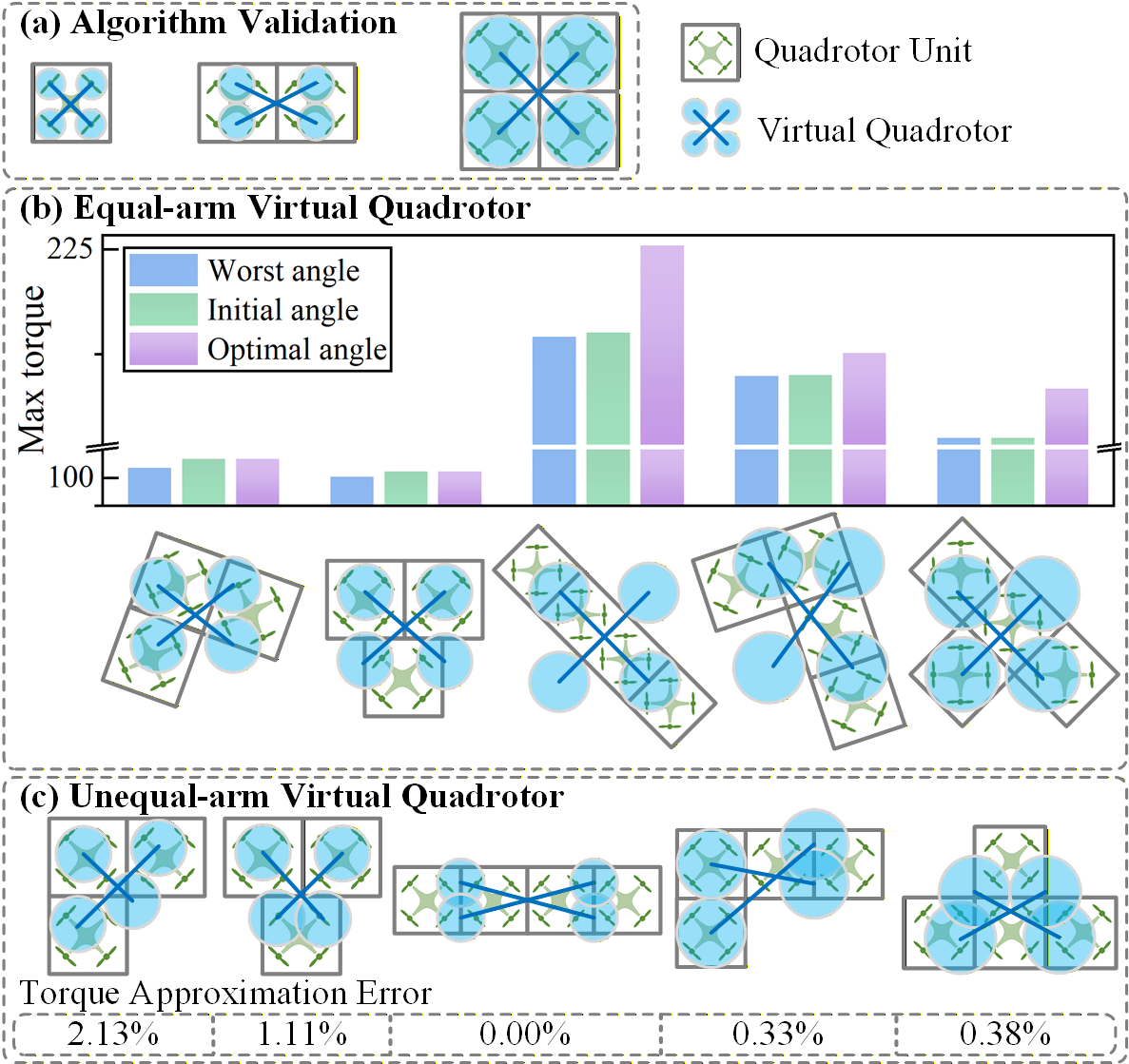}
    \vspace{-8mm}
    \caption{Model abstraction for various MARS configurations.
(a) Algorithm validation using virtual quadrotors formed by one, two, and four units.
(b) Equal-arm virtual quadrotor. MARS with the optimal yaw angle exhibits an improvement in maximum net torque over the baseline configurations. 
(c) Unequal-arm virtual quadrotor. Computed virtual representations of MARS configurations, illustrating unbalanced torque along the $x$- and $y$-axes. Torque approximation error of each model abstraction, computed using~Eq. \eqref{eq:approxi_error}.}
\label{fig:virtual quadrotor} 
\vspace{-7.5mm}
\end{figure}
\section{Simulation Evaluation}
\label{sec:simulation}
This section evaluates both the equal-arm and unequal-arm model abstractions, and then examines the MARS trajectory tracking performance under our APC framework.
\subsection{Model Abstraction of MARS}
\subsubsection{Validation on Symmetric Structures}
Both the equal-arm and unequal-arm model abstractions are validated on one-, two-, and four-unit symmetric MARS configurations. As shown in Fig.~\ref{fig:virtual quadrotor}(a), gray squares with green drones denote physical units, while blue propellers represent the virtual quadrotor. For the single-unit case, the virtual quadrotor obtained from Algorithms~\ref{alg:StandardVirtual} and~\ref{alg:FlexibleVirtual} coincides with the physical unit. For the two-unit case, a pair of rotors on one side of the virtual quadrotor corresponds to one unit. For the four-unit case, the virtual quadrotors returned by both algorithms also match expectations, where each unit is treated as one rotor of the virtual quadrotor.
\subsubsection{Equal-arm Model Abstraction}
As shown in Fig.~\ref{fig:virtual quadrotor}(b), the equal-arm $\times$-shaped virtual quadrotor is applied to more complex MARS configurations. Previous studies allocate rotor forces without adjusting the yaw angle of MARS, 
namely simply setting $\theta_z = 0$ in \eqref{eq:torque_theta_z}, 
which does not fully exploit the maximum net torque that MARS can generate \cite{Modquad,ModQuad-Vi,Modquad-gripper}.
In contrast, by solving the optimizing problem \eqref{eq:op_orientation}, we can not only exploit the maximum net torque, but also balance the torque between the $x$- and $y$-axes.
\begin{figure}[!t]
    \centering
    \includegraphics[width= 0.4 \textwidth]{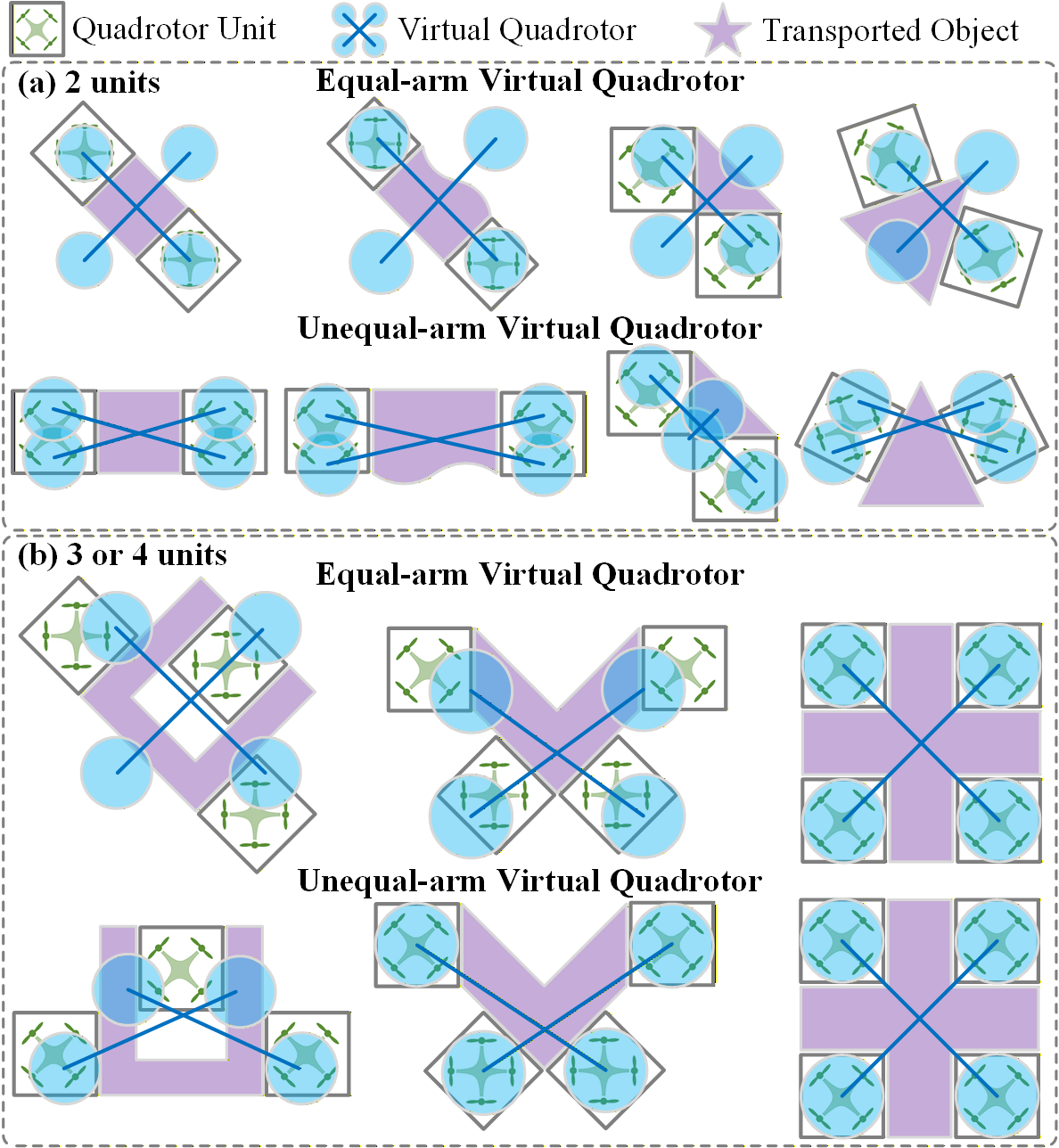}
    \vspace{-3mm}
    \caption{ Model abstraction  for arbitrary-shaped transportation using MARS. Unequal- and equal-arm quadrotor formations with (a) 2 units and (b) 3 or 4 units, illustrating object transport.}
    \label{fig:virtual quadrotor with object} 
    \vspace{-7mm}
\end{figure}
In the comparison, we compute the maximum achievable total net torque for $\theta_z = 0$, $\theta_z = \theta_z^{*}$, and $\theta_z = \theta_z^{\text{worst}}$ under different MARS configurations, where $\theta_z^{*}$ is solved from \eqref{eq:op_orientation}, and $\theta_z^{\text{worst}}$ yields the minimum net torque.
The comparison results are visualized in the bar chart in Fig.~\ref{fig:virtual quadrotor}(b), showing that our equal-arm model abstraction yields up to $15.94\%$ (average $7.63\%$) improvement in maximum net torque compared to the non-rotated case.
\begin{figure*}[!t]
\centering
    \includegraphics[width = 18 cm]{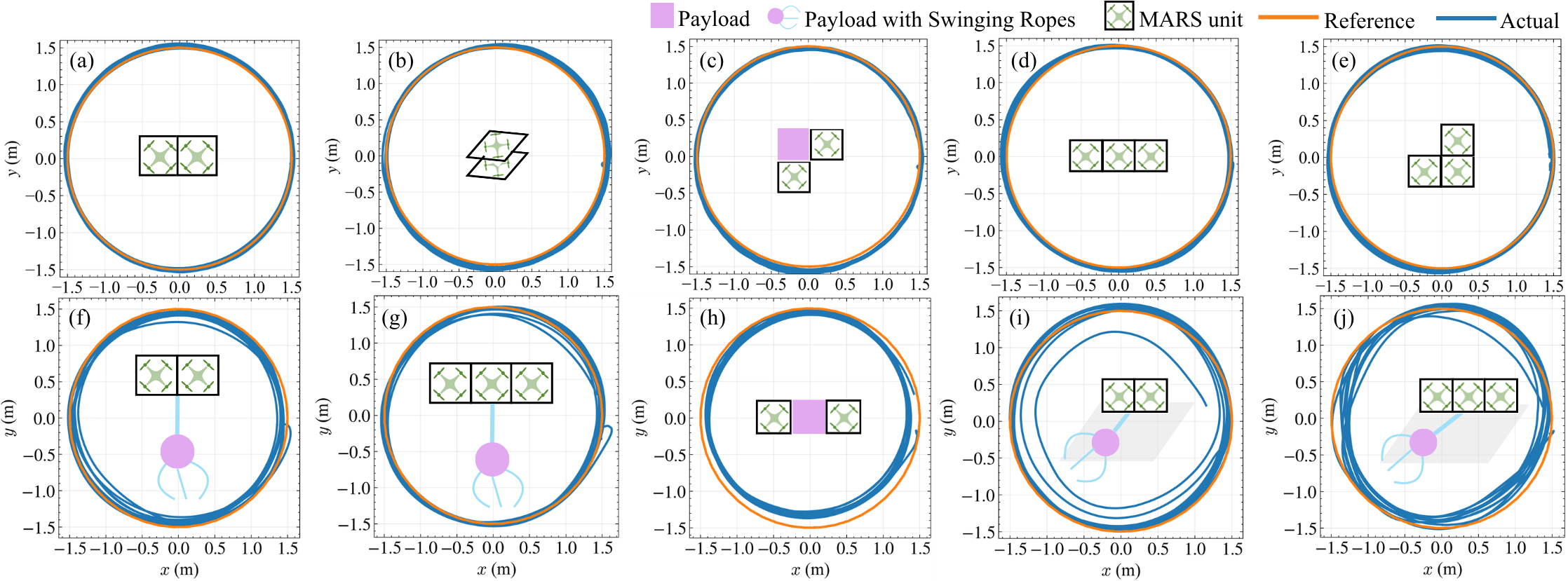}
\vspace{-4mm}
\caption{Simulation of trajectory tracking and object transport. (a)–(e) Trajectory tracking over ten consecutive laps along a circular path under different configurations. (f)–(j) Circular trajectory tracking on a 1.5 m radius path at 2 m/s with a 1.5 kg payload under different task settings.
}
\label{fig:sim_circular_tracking}
\vspace{-7mm}
\end{figure*}
\subsubsection{Unequal-arm Model Abstraction}
The unequal-arm virtual quadrotor shares the same coordinate frame as the MARS, with aligned yaw. Unlike the equal-arm model, it allows arbitrary rotor placement, where the geometry directly reflects control authority along the $x$- and $y$-axes (Fig.~\ref{fig:virtual quadrotor}(c)). For example, the third configuration has a longer arm (higher control authority) in the $y$-direction but a shorter arm (lower control authority) in the $x$-direction. This model abstraction is well-suited for transportation tasks in which yaw remains constant during flight, and can also guide the optimal unit placement for the MARS. Compared with the equal-arm model, which does not consider whether the MARS can actually generate the torque required by the virtual quadrotor, the unequal-arm model abstraction explicitly ensures that the control input of the virtual quadrotor can be realized by the MARS. However, the optimization problem \eqref{eq:flex_op_plbm} does not constrain the net torque of the virtual quadrotor to match that of the MARS; thus, torque loss arises, as quantified in \eqref{eq:approxi_error} and shown in Fig.~\ref{fig:virtual quadrotor}(c).
\subsubsection{Model Abstraction with Payload}
We extend the model abstraction methods to MARS transporting arbitrarily shaped payloads. As shown in Fig.~\ref{fig:virtual quadrotor with object}, the payload (purple) is rigidly connected to the MARS. Equal- and unequal-arm virtual quadrotors are constructed for formations of two to four units.
Unlike prior multi-quadrotor payload transport methods that rely on suspended cables~\cite{ zeng2025decentralized,sun2025agile, agarwal2025decentralized, wang2025safe}, our approach rigidly attaches the payload to the MARS. This enables precise modeling of the whole transportation system. Moreover, the model abstraction seamlessly supports planning and control without considering the payload’s influence, providing a novel and effective paradigm for stable and accurate aerial transportation of irregularly shaped objects. Moreover, the unequal-arm virtual quadrotor facilitates optimal unit placement around the payload for improved load distribution.
\subsubsection{Summary}
By constructing these equal- and unequal-arm model abstractions, MARS can be represented as a single virtual quadrotor, eliminating the need for manually designed coordination rules used in prior work~\cite{Modquad,ModQuad-Vi}. This virtual quadrotor allows any model-based controller, such as geometric control, APC, or model reference adaptive control, to seamlessly control the entire system effectively.
\subsection{Simulation}
The APC in \eqref{eq:APC}, using the virtual quadrotor as its predictive model, is employed to achieve agile trajectory tracking in this section. The APC was tested in the Gazebo simulation with PX4 SITL, and the control framework is shown in Fig.~\ref{fig:APC}. The drone \textit{Iris} (1.5 kg) was used as the unit model for MARS. The controller operating at 500Hz, publishes control outputs via a Robot Operating System (ROS) topic. The controller implementation follows previous work~\cite{falanga2018paMPC, MPC-li2023nonlinear }.
\begin{table}[t]
\caption{Trajectory Tracking Error in Fig.\ref{fig:sim_circular_tracking}}
\label{tab:simulation_tracking_error}
\vspace{-2mm}
\centering
\scriptsize
\renewcommand{\arraystretch}{1.3}
\begin{tabular}{@{}ccccccc@{}} 
\toprule
(a) & (b) & (c) & (d) & (e) & Average \\
\midrule
0.0261 & 0.0364 & 0.0385
& 0.1264 & 0.1329 & 0.0721 \\
\bottomrule
\end{tabular}
\vspace{-4mm}
\end{table}
\begin{table}[t]
\setlength{\tabcolsep}{4pt}
\caption{Tracking error (m) under different velocities, payload weights, and configurations.}
\label{tab:sim_payload_tracking}
\centering
\vspace{-2mm}
\scriptsize
\renewcommand{\arraystretch}{1.2}
\begin{tabular}{cccccccc}
\hline
\multirow{2}{*}{\textbf{Config}} &
\multirow{2}{*}{\textbf{Method}} &
\multirow{2}{*}{\makecell{\textbf{Vel}\\\textbf{(m/s)}}} &
\multicolumn{3}{c}{\textbf{Payload (kg)}} &
\multirow{2}{*}{\makecell{\textbf{Method}\\\textbf{Avg}}} \\
\cline{4-6}
 &  &  & \textbf{0.5} & \textbf{1.0} & \textbf{1.5} &  \\
\hline

\multirow{3}{*}{\textbf{2$\times$1}}
& \multirow{1}{*}{Modquad\cite{Modquad-gripper}}
& 1 / 2 & * & * & * & \multirow{1}{*}{--} \\
\cline{2-6}
& \multirow{2}{*}{\makecell{Ours: APC}} 
& 1 & 0.1349 & 0.1599 & 0.1327 & \multirow{2}{*}{0.1815} \\
&  & 2 & 0.2134 & 0.2338 & 0.2142 &  \\
\hline

\multirow{3}{*}{\textbf{3$\times$1}}
& \multirow{1}{*}{Modquad\cite{Modquad-gripper}}
& 1 / 2 & * & * & * & \multirow{1}{*}{--} \\
\cline{2-6}
& \multirow{2}{*}{\makecell{Ours: APC}} 
& 1 & 0.1960 & 0.1111 & 0.1436 & \multirow{2}{*}{0.1619} \\
&  & 2 & 0.1848 & 0.1493 & 0.1864 &  \\
\hline

\multirow{3}{*}{\makecell{\textbf{2$\times$1}\\\textbf{Attached}}}
& \multirow{1}{*}{Modquad\cite{Modquad-gripper}}
& 1 / 2 & * & * & * & \multirow{1}{*}{--} \\
\cline{2-6}
& \multirow{2}{*}{\makecell{Ours: APC}} 
& 1 & 0.1702 & 0.1932 & 0.1420 & \multirow{2}{*}{0.1631} \\
&  & 2 & 0.1579 & 0.1660 & 0.1492 &  \\
\hline

\multirow{3}{*}{\makecell{\textbf{2$\times$1}\\\textbf{Dragging}}}
& \multirow{1}{*}{Modquad\cite{Modquad-gripper}}
& 1 / 2 & * & * & * & \multirow{1}{*}{--} \\
\cline{2-6}
& \multirow{2}{*}{\makecell{Ours: APC}} 
& 1 & 0.1985 & 0.2752 & 0.4647 & \multirow{2}{*}{0.6577} \\
&  & 2 & 0.2837 & 0.8447 & $1.8791^*$ &  \\
\hline

\multirow{3}{*}{\makecell{\textbf{3$\times$1}\\\textbf{Dragging}}}
& \multirow{1}{*}{Modquad\cite{Modquad-gripper}}
& 1 / 2 & * & * & * & \multirow{1}{*}{--} \\
\cline{2-6}
& \multirow{2}{*}{\makecell{Ours: APC}} 
& 1 & 0.1735 & 0.2358 & 0.3043 & \multirow{2}{*}{0.2670} \\
&  & 2 & 0.2430 & 0.2744 & 0.3708 &  \\
\hline
\end{tabular}\\
\scriptsize\textit{* indicates that the system is able to take off but crashes during tracking.}
\vspace{-6mm}
\end{table}
\subsubsection{Trajectory Tracking}
\begin{table}[t]
\caption{Comparison of Agile Transport}
\vspace{-2mm}
\label{tab:Agile Transport}
\centering
\scriptsize
\renewcommand{\arraystretch}{1.3}
\begin{tabular}{@{}lccccc@{}}
\toprule
\makecell{\textbf{Payload}\\\textbf{(kg)}} 
& \makecell{\textbf{Method}} 
& \makecell{\textbf{XY}\\\textbf{Error (m)}} 
& \makecell{\textbf{Max Height}\\\textbf{Error (m)}} 
& \makecell{\textbf{Stability}} \\
\midrule

\multirow{2}{*}{0.5} 
& Modquad~\cite{Modquad,Modquad-gripper} & N/A & N/A & N/A \\
& Ours: APC & 0.2982 & 0.5294 & Stable \\
\midrule

\multirow{2}{*}{1.0} 
& Modquad~\cite{Modquad,Modquad-gripper} & N/A & N/A & N/A \\
& Ours: APC & 0.3788 & 2.5023 & Unstable \\
\bottomrule
\end{tabular}
\\
\scriptsize \textit{N/A indicates not applicable to this task.}
\vspace{-5mm}
\end{table}
We first evaluate the proposed model abstraction and APC on a circular trajectory tracking task of ten consecutive laps, with the radius of $1.5 \, \mathrm{m}$ and the desired velocity of $1 \, \mathrm{m/s}$.
Figs.~\ref{fig:sim_circular_tracking}(a)–(e) compare the reference and actual trajectories of MARS under different configurations. 
The results show that, with the APC, MARS tracks the desired trajectory with small errors, even when carrying a payload, thereby demonstrating the effectiveness and applicability of the proposed framework.  The mean absolute position error is summarized in Table~\ref{tab:simulation_tracking_error}, with an average of $0.0721\,\mathrm{m}$ over ten laps. 
Considering the MARS size up to $1.95\,\mathrm{m}$ (triple the side length $0.65\,\mathrm{m}$ of the \textit{Iris} model), this error level is negligible relative to vehicle scale. 
\subsubsection{Agile Transport}
We compute the metrics in Fig.~\ref{fig:sim_circular_tracking}(f)–(j), summarized in Table~\ref{tab:sim_payload_tracking}. Both Modquad~\cite{Modquad-gripper} and our method do not explicitly model the payload; however, the prior approach fails under these conditions. In contrast, our method maintains robust performance across configurations, velocities, and payload masses despite unmodeled dynamics. This holds for both suspended payloads and more challenging payload-dragging scenarios with ground contact disturbances. We further evaluate control performance in agile flight~\cite{geles2024demonstrating,romero2025actor,hanover2024autonomous}. Experiments use $0.5~\mathrm{kg}$ and $1.0~\mathrm{kg}$ suspended payloads with $450~\mathrm{g}$ ropes. Aggressive reference trajectories are generated using the minimum-snap method~\cite{richter2016polynomial}, with a desired velocity of $4~\mathrm{m/s}$ and a target pitch angle of $40^{\circ}$. Our method achieves this pitch angle in both cases. Table~\ref{tab:Agile Transport} reports the tracking errors, which increase with payload mass. Overall, the results show that the proposed APC maintains stability under severe payload-induced unmodeled dynamics.
\subsubsection{Ablation Discussion}
\begin{table}[t]
\caption{Tracking Error (m) under Different Ablation Settings}
\label{tab:ablation}
\centering
\vspace{-2mm}
\scriptsize
\setlength{\tabcolsep}{3pt}
\begin{tabular}{@{}lcccccc@{}}
\toprule
\multirow{3}{*}{\makecell{\textbf{Model + Controller}}}
& \multirow{3}{*}{\makecell{\textbf{Allocation}}}
& \multicolumn{3}{c}{\textbf{Task}}
& \multirow{3}{*}{\makecell{\textbf{Average}}} \\
\cmidrule(lr){3-5}
& & 
\makecell{\textbf{Tracking}}
& \makecell{\textbf{Payload}}
& \makecell{\textbf{Dragging}} & \\
\midrule
Simplified model + PID & Rule-based & $0.6661$ & $1.9003^*$ & $1.0624^*$ & $1.2096$\\
Ours: APC& Rule-based & $0.1764$ & $0.7313^*$ & $1.3434^*$ & $0.7504$ \\ 
\rowcolor{gray!10}
Ours: APC& Dynamic & $\boldsymbol{0.0261}$ & $\boldsymbol{0.2142}$ & $\boldsymbol{0.4647}$ & $\boldsymbol{0.2350}$\\
\bottomrule
\end{tabular}\\
\scriptsize\textit{* indicates collisions or crashes during this task.}
\vspace{-6mm}
\end{table}
Finally, to evaluate each module in the proposed framework, we conduct an ablation study (Table~\ref{tab:ablation}). The baseline simplified model with PID and rule-based allocation performs worst, especially under payload and dragging conditions where failures occur. Replacing it with the proposed virtual quadrotor and APC improves tracking, but instability remains with rule-based allocation. In contrast, dynamic allocation achieves the lowest errors across all tasks, eliminates failures, and reduces the average error by up to $80.6\%$ to 0.2350 m. These results confirm that the virtual quadrotor abstraction and dynamic allocation are essential for robust, accurate performance, avoiding oscillations caused by simplified models and discontinuous control in prior approaches~\cite{Modquad-gripper}.
\begin{figure*}[!t]
\centering
    \includegraphics[width = 18 cm]{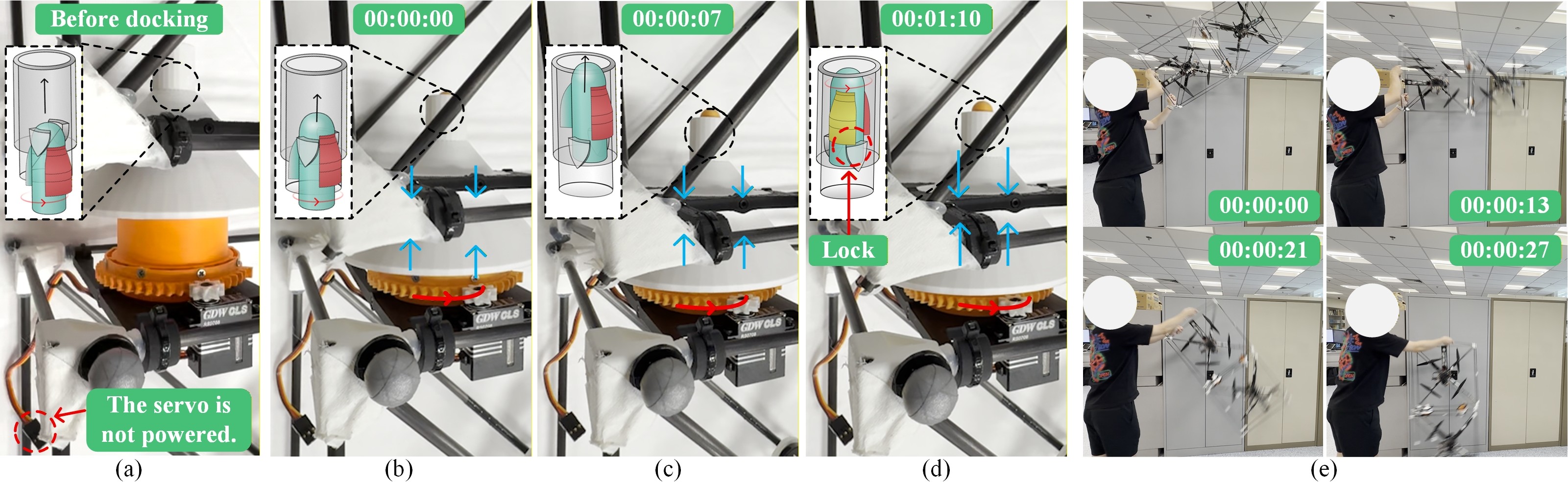}
\vspace{-4mm}
\caption{Passive docking and lock sequences. The timer format is \textit{min:sec:ms}.
 (a) Before docking: the servo is not powered, and the mechanism is in the standby state. (b) Initial docking: the male and female components begin to engage under the guidance of magnetic attraction. (c) The male docking mechanism is fully inserted into the female component and begins to lock. (d) The male component passively rotates due to the asymmetric magnetic field, reaching the designed locking position. (e) A $135^\circ$ pendulum test demonstrates the strong mechanical strength of the docked connection.}
\label{fig:docking and locking}
\vspace{-4mm}
\end{figure*}
\begin{figure*}[!t]
\centering
    \includegraphics[width = 18 cm]{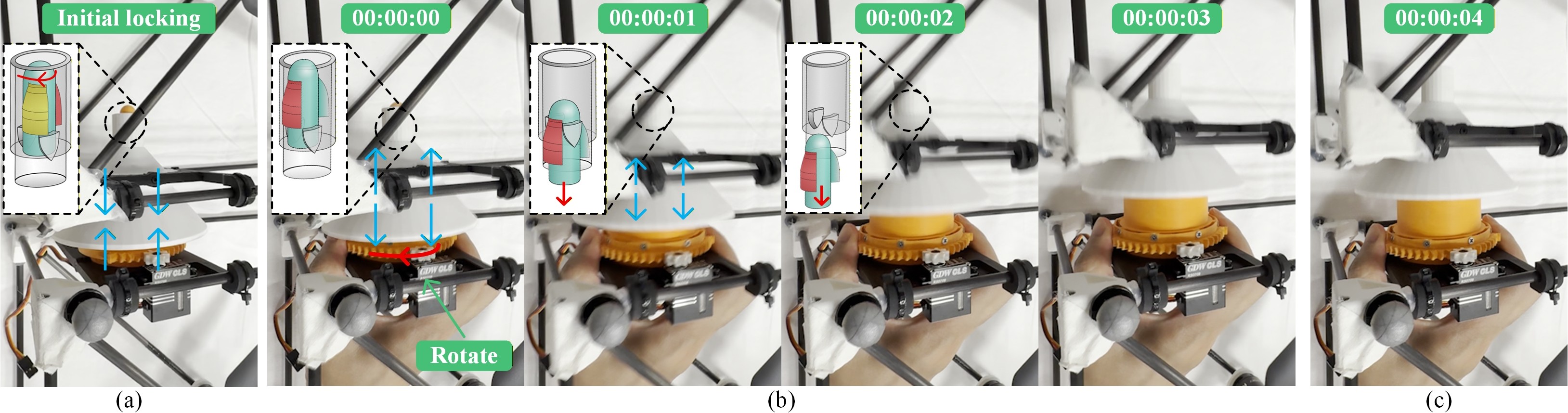}
\vspace{-4mm}
\caption{Manually triggered active separation rotates the male component to repel the connected module. (a) Initial locking state. (b) At 00:00:00, the male component of the rotational docking mechanism is manually rotated to unlock the docking-separation interface. The repulsive force from embedded magnets assists the separation process. (c) Fully separated state at 00:00:04.}
\label{fig:manually separation}
\vspace{-6mm}
\end{figure*}
\subsubsection{Summary}
With our framework, MARS performs as a single quadrotor without relying on manually designed, task-specific allocation rules adopted in previous works~\cite{Modquad,ModQuad-Vi,saldana2019design,Modquad-dof}. 
The simulation results of trajectory tracking and object transport highlight the generalization capability of the proposed model abstraction and controller, providing a unified and scalable framework for accurate tracking across diverse MARS configurations as well as high-performance agile flight.
\section{Real-World Experiments}
\label{sec:real-world}
\begin{table}[t]
\caption{Summary of Experimental Tasks Performed by MARS}
\label{tab:comparison_demos}
\centering
\vspace{-2mm}
{\fontsize{6pt}{8pt}\selectfont
\renewcommand{\arraystretch}{1.3}
\begin{tabular}{@{}lcccccc@{}}
\toprule
\makecell {\textbf{Method}} 
& \makecell  {\textbf{Docking}}
& \makecell  {\textbf{Separation}}
& \makecell  {\textbf{Agile} \\ \textbf{tracking}}
&  \makecell{\textbf{Agile} \\ \textbf{flight}} 
&  \makecell{\textbf{Collision} \\ \textbf{free}}
&  \makecell{\textbf{Agile} \\ \textbf{transport}}
\\
\midrule
Modquad ~\cite{Modquad} & \ding{51}  & \ding{55} & \ding{55}  & \ding{55}  & \ding{55} & \ding{55} \\

Modquad-Vi ~\cite{ModQuad-Vi} & \ding{51}  & \ding{55} & \ding{55}  & \ding{55}  & \ding{55} & \ding{55}\\

Ref ~\cite{saldana2019design} & \ding{55}  & \ding{51}& \ding{55} & \ding{55}  & \ding{55}  & \ding{55}\\

Ref ~\cite{RAL-zhang2025design} & \ding{51}  & \ding{51}& \ding{55} & \ding{55}  & \ding{55}  & \ding{55}\\
\rowcolor{gray!10}
\textbf{Ours: APC} & \ding{51} & \ding{51} & \ding{51}  & \ding{51}  & \ding{51} & \ding{51}\\
\bottomrule
\end{tabular}
}
\vspace{-8mm}
\end{table}
In this section, we first conduct ground tests of our integrated docking mechanism, followed by mid-air docking and separation experiments.
We then perform a series of experiments, including trajectory tracking, agile flight, collision-free navigation, and agile transport, as summarized in Table~\ref{tab:comparison_demos}.
\subsection{Integrated Docking-Locking-Separation Mechanism}
\subsubsection{Self-Correction Docking}
During docking, due to various disturbances, the allowable docking position range may exceed the position control error. Our tests indicate that the docking mechanism can tolerate $x$- and $y$-axis position errors up to $\pm 35$ mm.
With the aid of the attractive magnetic force (approximately 100 N), the docking mechanism accommodates rotational errors of up to $12.5^\circ$ in roll, $15.5^\circ$ in yaw, and $10^\circ$ in pitch. 
Since the system is underactuated, attitude maneuvers are required for docking; in experiments, roll and pitch deviations of up to $3^\circ$ were observed (Fig.~\ref{fig:flight_docking}), confirming that the mechanism tolerance is sufficient.


\subsubsection{Detection-Free Passive Lock}
\begin{figure*}[!t]
\centering
    \includegraphics[width = 18 cm]{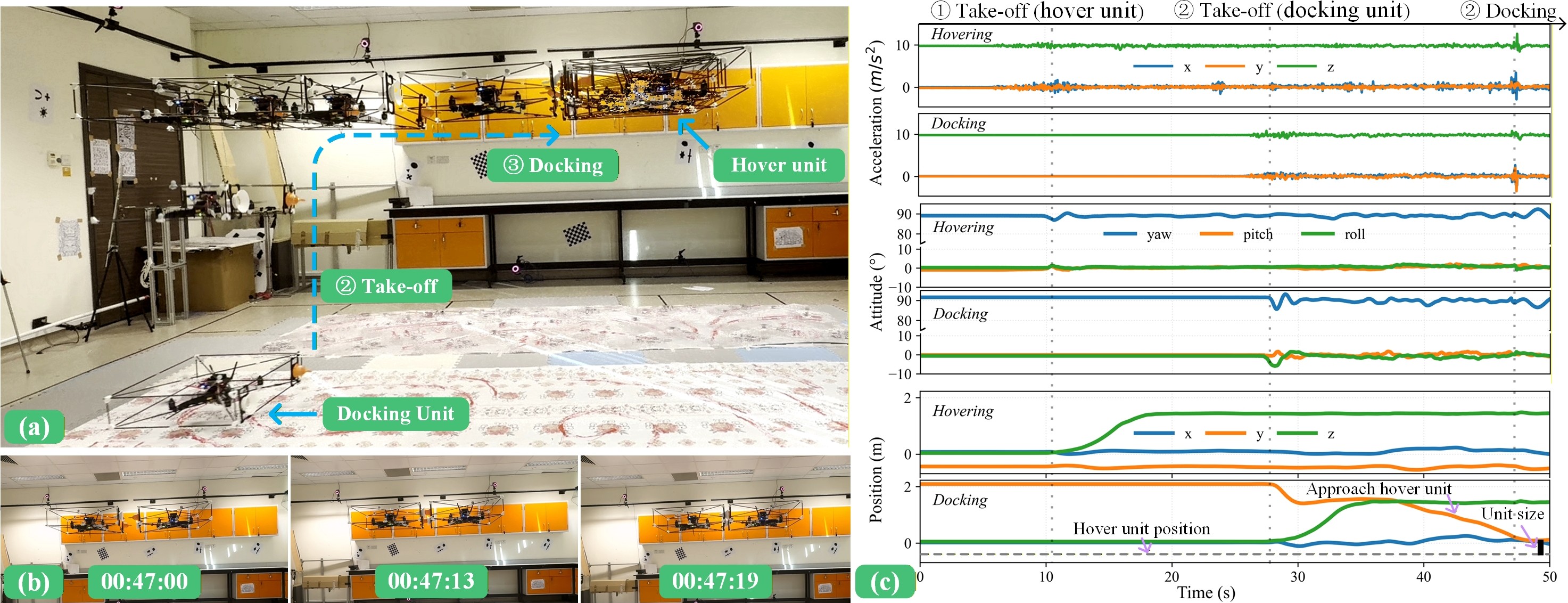}
\vspace{-4mm}
\caption{Mid-air docking demonstration. The timer format is \textit{min:sec:ms}.
(a) The hover unit maintains a stable position while the docking unit takes off, approaches, and actively adjusts its trajectory to achieve precise mid-air docking. (b) The docking process starts at 00:47:00 and successfully ends at 00:47:19.
(c) Time-series plots of acceleration, attitude angles, and position for both units, showing active trajectory correction of the docking unit, and the docking collision, illustrating the stability of the proposed mid-air docking approach.}
\label{fig:flight_docking}
\vspace{-5mm}
\end{figure*}

Mid-air locking is essential because pure magnetic connections \cite{Modquad,ModQuad-Vi} might break under disturbances or unbalanced payloads. 
Latest studies~\cite{Tokyo_sugihara2024beatle} incorporated mechanical locking to improve load capacity (Table~\ref{tab:docking mechanism comparison}, fifth row). 
However, these designs require detection of the precise docking moment; if locking is triggered prematurely, the docking fails. Accurate detection is difficult because docking events often resemble collisions, as evidenced by our docking experiments (Fig.~\ref{fig:flight_docking}(c)). Our detection-free docking mechanism addresses these limitations. As shown in Fig.~\ref{fig:docking and locking}, when the male mechanism approaches the female component (a), magnetic attraction guides the male head into the female tail (b). Once fully inserted (c), the male component passively rotates under the influence of an asymmetric magnetic field, reaching the designed locking position within 1.003~s, as shown in Fig.~\ref{fig:docking and locking}(d)). The connection strength is further evaluated using the pendulum test (Fig.~\ref{fig:docking and locking}(e)), demonstrating high locking strength. 
In summary, our locking mechanism eliminates the need for a servo motor to actuate rotation, removes the docking detection requirement, and enables automatic locking immediately upon full insertion, highlighting the simplicity and effectiveness of the design.
\subsubsection{Magnetic Repulsion-based Separation}
\begin{table}[t]
\caption{Comparison of docking methods}
\label{tab:comparison_docking}
\centering
\vspace{-2mm}
\footnotesize
\scriptsize
\renewcommand{\arraystretch}{1.3}
\begin{tabular}{@{}lcccccc@{}}
\toprule
\makecell {\textbf{Method}} 
& \makecell {\textbf{Docking} \\ \textbf{Mechanism}} 
&  \makecell{\textbf{Max x/y Acceleration}\\ ($\text{m/s}^2$)} 
&  \makecell{\textbf{Normalized} \\ \textbf{ height error}}\\
\midrule

Modquad ~\cite{Modquad} & Magnet  & Not reported & \makecell{2.02 = \\0.097 / frame size} \\

ModQuad-Vi ~\cite{ModQuad-Vi} & Magnet  & 4.9  & \makecell{{5.15 =}\\ {0.577 / frame size}}\\
\rowcolor{gray!10}
\textbf{Ours} & \textbf{Mode shift} & \textbf{3.47} & \makecell{\textbf{0.28 =
} \\ {\textbf{0.042} / frame size} } \\ 
\bottomrule
\end{tabular}
\vspace{-7mm}
\end{table}
Previous studies in \cite{saldana2019design} proposed to use separation torque to actively detach two connected units (Table~\ref{tab:docking mechanism comparison}, row~3), but the resulting attitude change may introduce safety risks. 
Subsequent work in \cite{Tokyo_sugihara2024beatle} employed a servo to reverse magnet polarity and a servo-driven slider-crank mechanism to retract a locking stick (Table~\ref{tab:docking mechanism comparison}, row~5). 
However, these approaches are complex, increase system weight, and are difficult to implement on all sides of the drone (e.g., four mechanisms per unit).

To ensure complete detachment and prevent the male component from jamming due to pose disturbances [cf. Fig.~\ref{fig:mechanism_detail}(d)], our design uses a servo motor to rotate the male component and reverse the magnet polarity, thereby generating a repulsive force to separation without attitude maneuver. 
Fig.~\ref{fig:manually separation} shows the separation sequence, where the docking mechanism is manually rotated to mimic servo actuation. Starting from the locked state, rotation of the male component begins at 00:00:00, and the two drones fully detach within 4 ms, demonstrating rapid separation capability.
\subsection{Flight Experiment Setup}
In the real-world experiments, we used a quadrotor with a 380-mm wheelbase and a carbon-fiber unit frame. An NVIDIA Jetson Orin module (CPU only, 2 GHz) was used as the onboard computer. The docking mechanism was mounted on one side of each quadrotor, and the total mass was approximately 1.55 kg. The PX4 firmware was modified to provide high-frequency inertial measurement unit (IMU) and position data. For the controller, the virtual control command $\boldsymbol{u}_\mathcal{V}$ is generated by the leader unit at approximately 500 Hz. This command is mapped to individual unit commands using \eqref{eq:op_allocate} and transmitted to other units over Wi-Fi with approximately 10 ms delay. The PX4 actuator topic is then used to directly send commands to the individual units in MARS. For scalability, the APC solve time without allocation remains $O(1)$, independent of the number of units, while control allocation is formulated as a QP and solved via qpOASES ($O(n^3)$, where $n$ is the number of units). For $n=50$, the solver takes $0.70~\mathrm{ms}$ on a $2~\mathrm{GHz}$ NVIDIA Jetson Orin CPU, demonstrating real-time feasibility for large-scale systems.
\subsection{Docking and Separation in the Mid-Air}
\subsubsection{Mid-Air Docking}
Fig.~\ref{fig:flight_docking}(a) illustrates the complete process of mid-air docking. Initially, the docking drone waits at the start position. Once the hovering unit reaches the designated target location, the docking drone takes off, approaches the hovering unit, and prepares to execute the docking operation at 00:47:00, successfully completing docking at 00:47:19.
Figure~\ref{fig:flight_docking}(c) presents the acceleration, attitude, and position of both drones over time during the docking process.
During docking, the $z$-axis acceleration of the hovering unit peaks at about 12.63 $\text{m/s}^2$ due to the collision, 
and is stabilized within 1s.
Similar transient variations are observed along the $x$- and $y$-axes, with maximum accelerations reaching 3.47 $\text{m/s}^2$ for both drones [cf. Table \ref{tab:comparison_docking}]. 
At the moment of docking, the vertical position of the MARS deviates from the desired value. Our docking mechanism and controller significantly reduce this deviation from 0.097 or 0.577~m to 0.042~m compared with~\cite{Modquad,ModQuad-Vi}, lowering the normalized height error from 5.15 or 2.02 to 0.28 and suppressing vertical oscillations. In summary, these results demonstrate that our design enables smooth docking with minimal attitude disturbance, confirming the effectiveness of the mechanism and control framework.
\subsubsection{Magnetic Repulsion-assisted Separation}
\begin{figure*}[!t]
\centering
    \includegraphics[width = 18 cm]{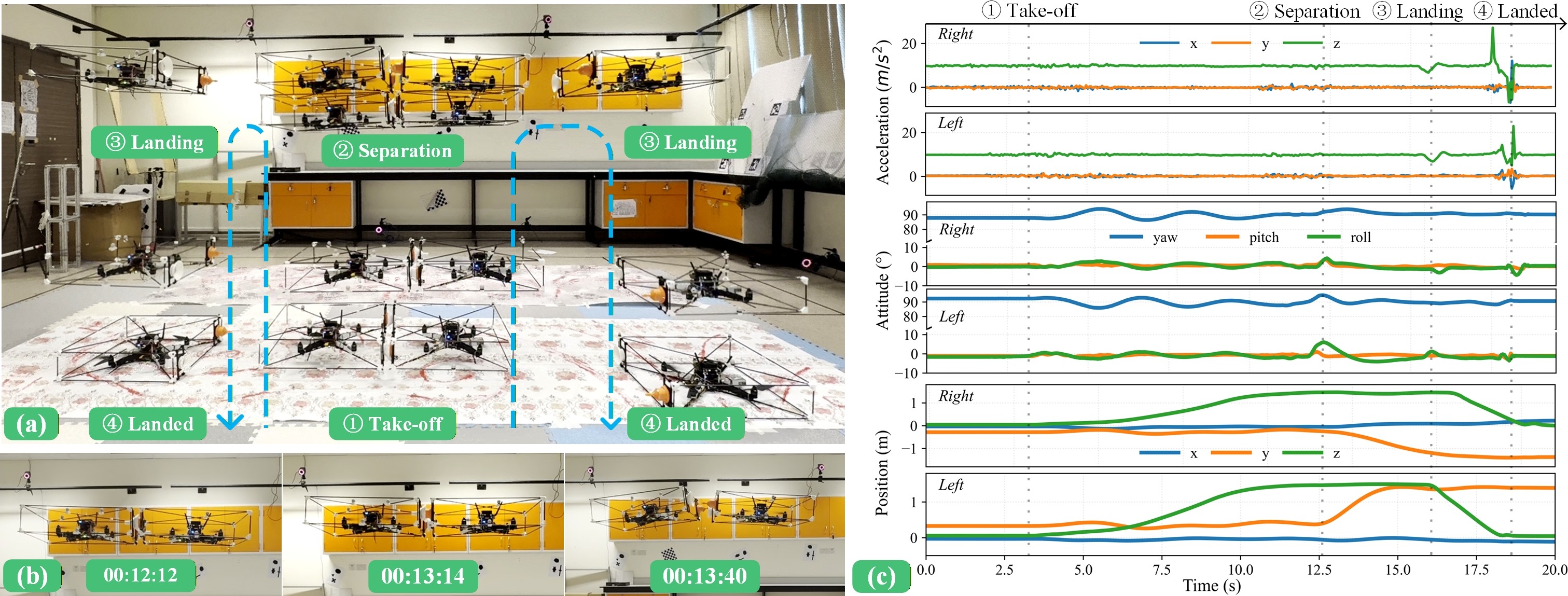}
\vspace{-4mm}
\caption{Mid-air separation demonstration. The timer format is \textit{min:sec:ms}.
(a) Two drones docked together take off simultaneously, perform a mid-air separation maneuver, and then independently land. (b) After switching to separation mode, the drones successfully disengage from each other during the time interval 00:12:12 to 00:13:40.
(c) Experimental data during the demonstration, showing the acceleration along x, y, z axes, attitude angles (pitch, roll, yaw), and position trajectories. The key events is marked on the plot, including take-off, mid-air separation, and landing states.}
\label{fig:flight_separation}
\vspace{-4mm}
\end{figure*}
\begin{figure*}[!t]
\centering
    \includegraphics[width = 17.5 cm]{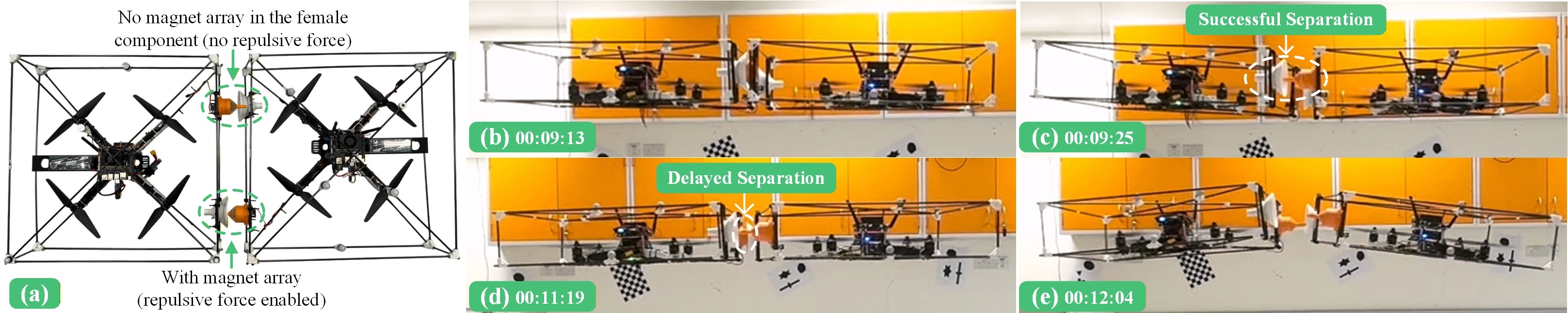}
\vspace{-4mm}
\caption{Separation comparison experiments. (a) One side enabled with repulsive force; the other disabled. (b) Initial position at 00:09:13. (c) The mechanism employing a repulsive force achieves successful separation at 00:09:25. (d) Without repulsive force, separation fails until 00:11:19, with a delay of approximately 2 s. (e) Successful separation occurs at 00:12:04.}
\label{fig:failed_flight_separation}
\vspace{-7mm}
\end{figure*}
For MARS units based on underactuated quadrotors~\cite{saldana2019design}, the system lacks sufficient actuation authority to directly generate separation torque. Instead, separation is achieved through attitude maneuvers. For example, \cite{saldana2019design} tilts the quadrotor to break the magnetic link, causing large attitude deviations (about $28^\circ$ in Table~\ref{tab:comparison_separation}) that degrade flight performance, especially with payloads. In contrast, our separation is driven by a 20 g mini servo motor that rotates the male docking mechanism from lock mode to separation mode (Fig.~\ref{fig:mechanism_detail}(c), Step 4–5). This design offers two advantages: (a) \textit{Improved stability}, since separation does not require attitude changes; (b) \textit{Higher success rate}, as the separation mode switches the magnetic force from attraction to repulsion, preventing re-locking with the female interface (Fig.~\ref{fig:mechanism_detail}(d)).
\begin{table}[t]
\caption{Comparison of separation methods}
\label{tab:comparison_separation}
\centering
\vspace{-2mm}
\footnotesize
\scriptsize
\renewcommand{\arraystretch}{1.3}
\begin{tabular}{@{}lcccccc@{}}
\toprule
\makecell {\textbf{Method}} 
& \makecell{\textbf{Separation}\\\textbf{method}} 
&  \makecell{\textbf{Maximum tilting}\\  \textbf{angle} ($^\circ$)} 
&  \makecell{\textbf{Separation}\\ \textbf{time} (s)} \\
\midrule
Ref.~\cite{saldana2019design} & Separation torque  & about 28 & 5 \\
Ref. \cite{zhang2024design} & Electromagnetic & \textbf{4} (Fully Actuated) & 3\\
Ours & Mode shift & 7.2 & 2.9\\
\rowcolor{gray!10}
\textbf{Ours} & \textbf{Mode shift} + \textbf{repulsive force} & 5.8 & \textbf{0.5}\\
\bottomrule
\end{tabular}
\footnotesize\textit{Separation time: the duration from separation initiation to stable flight.}
\vspace{-8mm}
\end{table}

Fig.~\ref{fig:flight_separation}(a) presents the aerial separation experiment. 
The top panel presents a composite image formed by key frames extracted from the experiment video. At first, the two drones remain docked in midair. Upon reaching the target location, the male docking mechanism switches from lock mode to separation mode at 00:13:14, activating a repulsive force that facilitates detachment. At 00:13:40, the position controller further drives the two drones apart. The Fig.~\ref{fig:flight_separation}(b) highlights these key timestamps, showing the male component (orange) separating from the female component (white) with the assistance of the repulsive force. Fig.~\ref{fig:flight_separation}(c) plots the time profiles of position, attitude, and acceleration during the separation process. The roll angle stays within $\pm6^\circ$, indicating that our active separation mechanism minimizes attitude deviation. The slight roll variations arise mainly from the position controller, which uses roll maneuvering to drive the two drones apart, rather than from any destabilizing interaction between the docking components. Additionally, the acceleration remains close to zero, and the position trajectories are smooth. In contrast, although prior work~\cite{saldana2019design} did not provide numerical separation curves, their videos show noticeable attitude oscillations and positional deviations in their separation. A quantitative comparison is provided in Table~\ref{tab:comparison_separation}: our method reduces the maximum tilting angle from roughly $28^\circ$ to $5.8^\circ$ (close to the $4^\circ$ achieved by fully actuated quadrotor \cite{zhang2024design}), and shortens the separation time from 5~s to 0.5~s, achieving both higher reliability and faster execution.
\begin{figure*}[!t]
\centering
    \includegraphics[width = 18 cm]{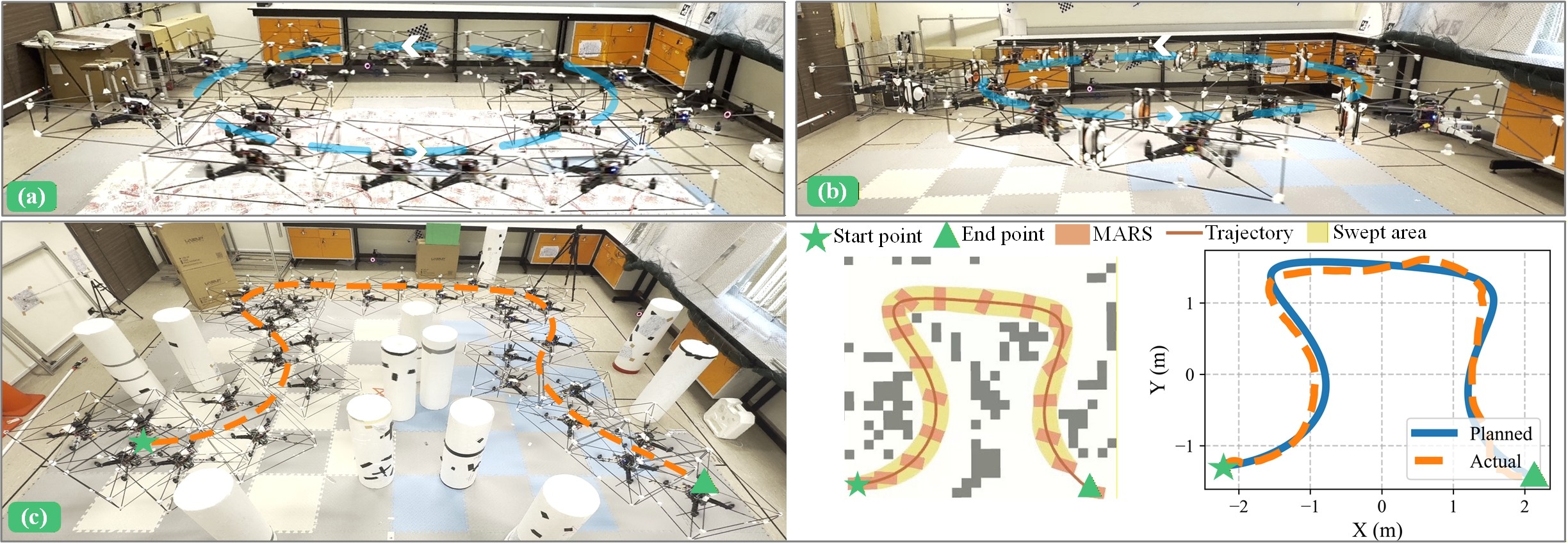}
\vspace{-4mm}
\caption{Composite images of trajectory tracking. (a) Trajectory tracking at a velocity of 1 $\text{m/s}$. (b) Circular trajectory tracking with the docking mechanism engaged and a desired yaw angle. (c) Collision-free navigation illustrating the robust tracking performance of MARS.}
\vspace{-4mm}
\label{fig:trajectory_tracking}
\end{figure*}
\begin{figure*}[!t]
\centering
    \includegraphics[width = 18 cm]{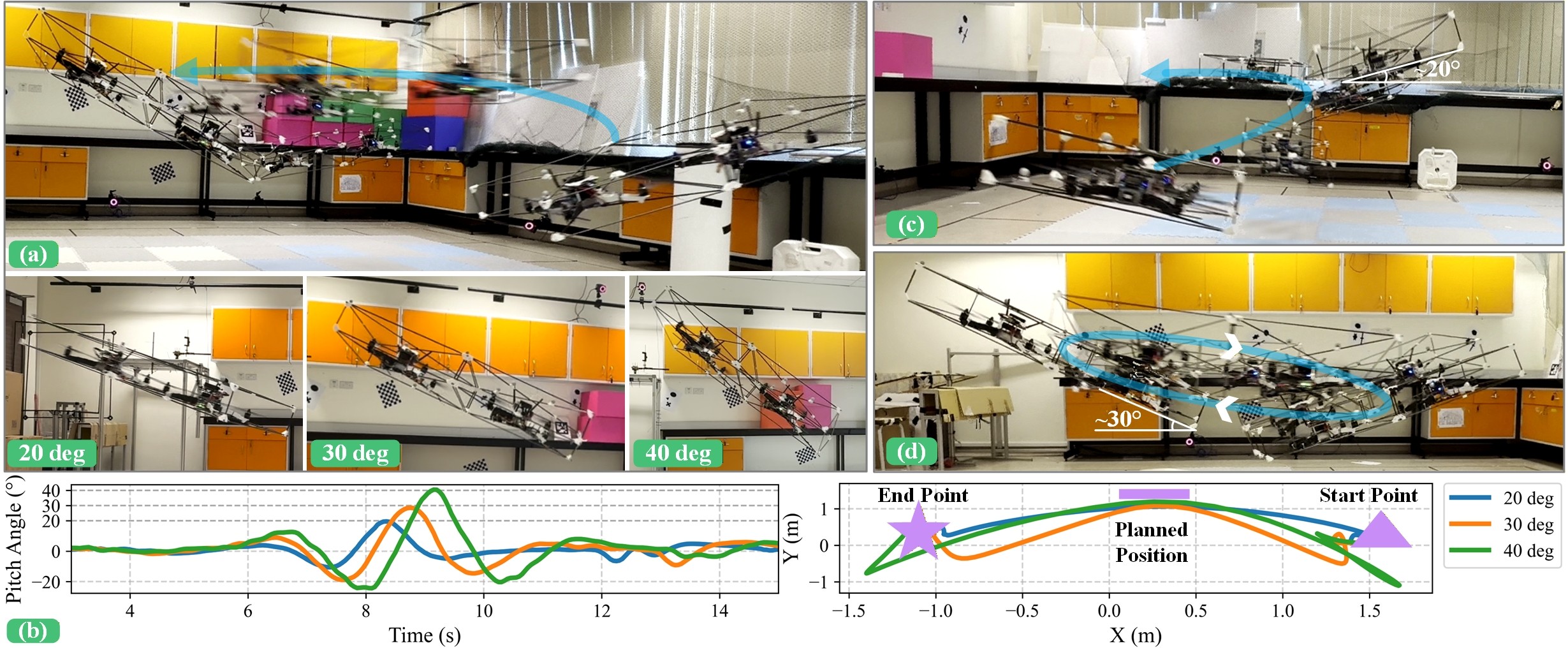}
\vspace{-4mm}
\caption{Agile trajectory tracking. (a) Pitch-angle responses along agile trajectories at $20^\circ$, $30^\circ$, and $40^\circ$ without retuning the APC parameters, demonstrating robustness to varying attitude demands. (b) Pitch-angle tracking in (a) was evaluated at the planned spatial path, further demonstrating the robustness of the agile flight controller. (c) Roll-angle tracking along an agile trajectory. (d) The MARS maintains speeds exceeding $2 ~\text{m/s}$ while following a $0.5 ~\text{m}$-radius circular path; the roll and pitch angles approach $30^\circ$, indicating high-agility performance. }
\label{fig:agile_trajectory}
\vspace{-6mm}
\end{figure*}

To further demonstrate the benefit of our proposed magnetic repulsion-assisted separation experimentally, we equipped one side of the docking mechanism with magnet arrays to generate the repulsive force, while leaving the opposite side without magnet. This asymmetric setup allows a direct comparison between the repulsive and non-repulsive cases. As shown in Fig.~\ref{fig:failed_flight_separation}, the separation sequence began at 00:09:13. With the repulsive force enabled, successful separation occurred at 00:09:25, whereas without it, components did not separate until 00:11:19 (an approximately 2 s delay). The cause of the failed or delayed separation in Fig.~\ref{fig:failed_flight_separation}(c)–(d) is explained in detail in Fig.~\ref{fig:mechanism_detail}(d), where uncertain contact conditions may cause the male connector to lock inside the female mechanism. These results highlight the effectiveness and robustness of our proposed repulsion–assisted separation mechanism.
\subsubsection{Compared with Previous Works}
\begin{figure}[!t]
\centering
    \includegraphics[width = 9 cm]{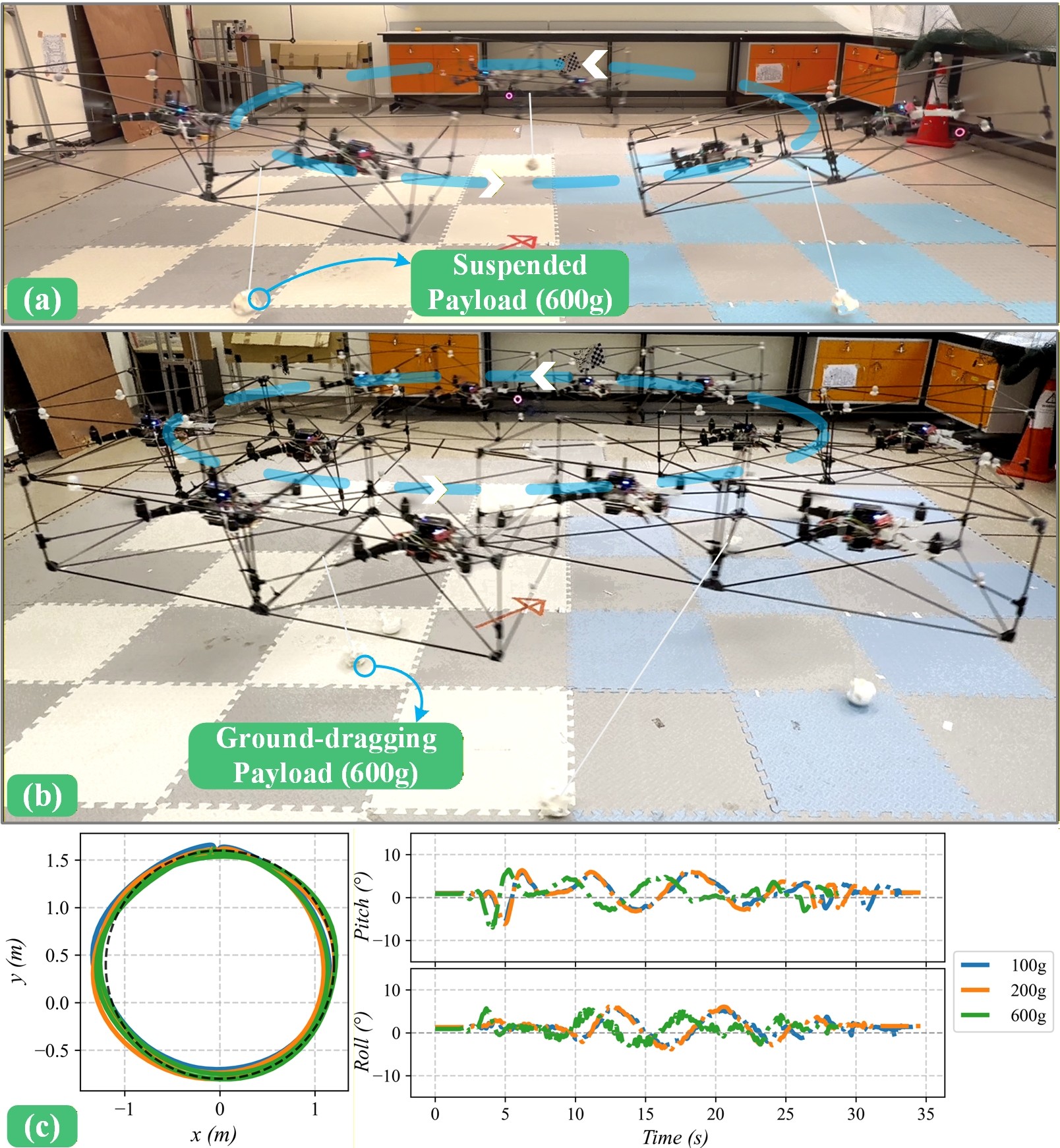}
\vspace{-8mm}
\caption{Object transport and dragging (600 g payload, 20\% of MARS weight).
(a) Circular tracking at 2 m/s demonstrates robust performance while carrying the payload.
(b) Payload dragging on the ground further demonstrates robustness under payload–ground contact disturbances. (c) Overlaid trajectories highlight accurate tracking performance.}
\label{fig:real_transport}
\vspace{-6mm}
\end{figure}
Our design provides the following key capabilities:
\textit{a) Integrated mechanism:}
Unlike previous work that designs one mechanism for docking~\cite{Modquad,ModQuad-Vi} and another for separation~\cite{saldana2019design}, our design integrates docking, locking, and separation into a single, simple, and compact structure driven by a micro-servo motor.
\textit{b) High-reliability docking:}
Prior mechanisms sometimes fail to dock under aerodynamic disturbances and, due to limited control accuracy~\cite{Modquad}. In contrast, our mechanism, together with the self-aligning geometric structure and magnetic attraction forces, achieves robust docking even under substantial approach misalignment. 
\textit{c) Passive detection-free locking:}
Previous works \cite{Modquad,ModQuad-Vi,saldana2019design} lack a dedicated locking mechanism, leaving assembled structures vulnerable to disturbances and contact forces. Reliable post-docking locking is therefore essential for maintaining integrity during aggressive maneuvers. However, active locking designs that rely on sensors and actuators struggle to reliably detect the brief moment of successful docking, as units can separate almost immediately due to impacts and aerodynamic disturbances. Our mechanism overcomes this limitation through fully passive, sensor-free locking that engages automatically upon contact, ensuring stable assembly under transient disturbances.
\textit{d) Smooth in-flight separation:}
Compared to prior work~\cite{saldana2019design}, where separation depends on a large attitude maneuver that introduces risk during flight, our design employs magnetic repulsion to initiate separation, requiring only a minimal attitude change and thereby improving stability and safety. In addition, our separation mechanism reduces actuation complexity and hardware count relative to the dual-servo, multi-link system in~\cite{Tokyo_sugihara2024beatle}, resulting in a more streamlined and reliable implementation.
\subsection{Trajectory Planning and Control}
Prior works achieved full 6-DoF control by using tilt-rotor quadrotor as the MARS unit, either through manually assembled tilt mechanisms~\cite{H-modquad,Tokyo_sugihara2023design} or by adding a servo motor to each rotor~\cite{Tokyo_sugihara2024beatle}. 
Although such designs enable straightforward trajectory tracking, the added actuators increase system cost and complexity.
\begin{figure}[!t]
\centering
    \includegraphics[width = 8.8 cm]{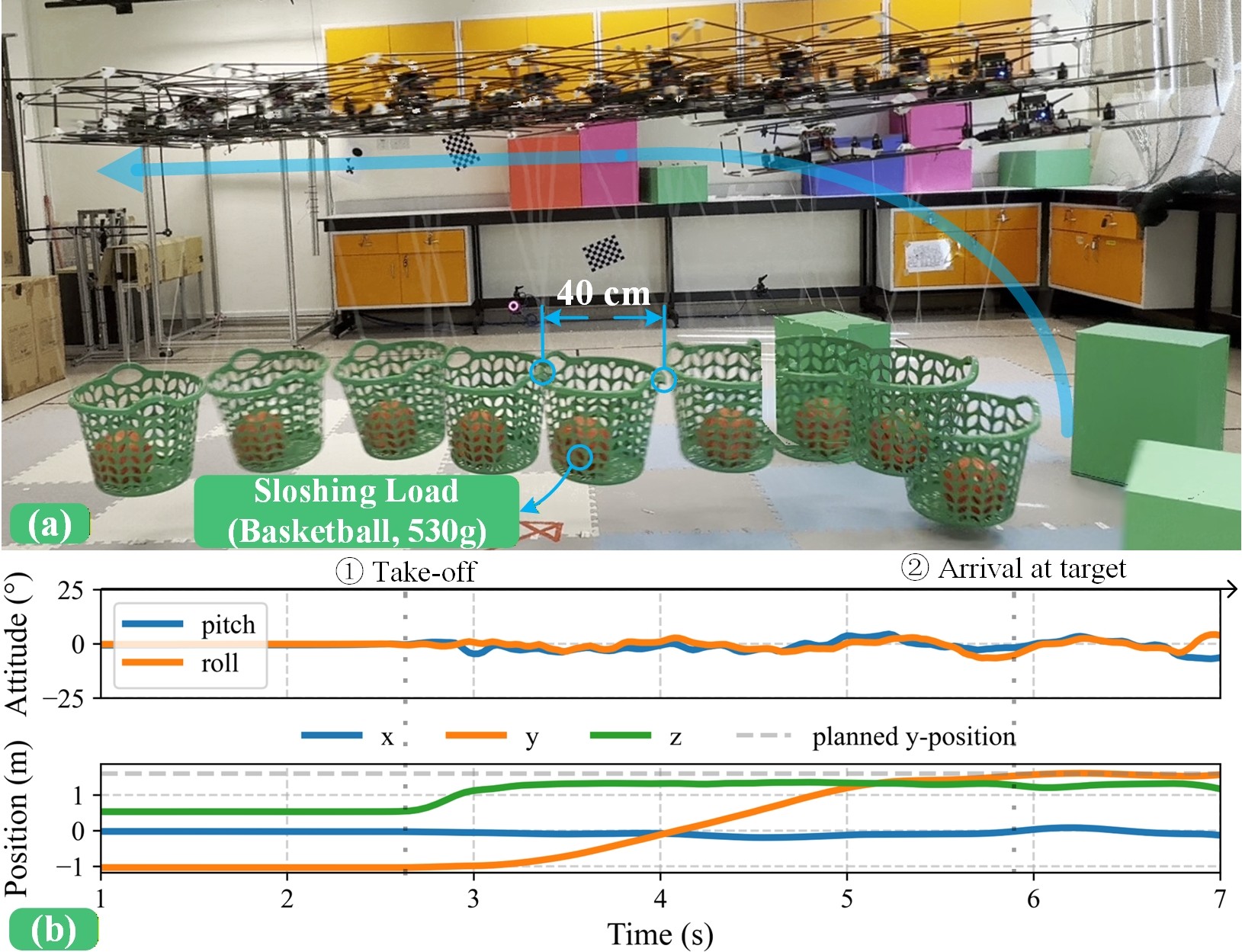}
\vspace{-8mm}
\caption{Object transportation with a sloshing load.
(a) Composite images show the basketball’s oscillatory motion during transport.
(b) Attitude error reflects sloshing effects; position tracking confirms controller robustness.
}
\label{fig:transport}
\vspace{-7.5mm}
\end{figure}
\begin{figure*}[!t]
\centering
    \includegraphics[width = 18 cm]{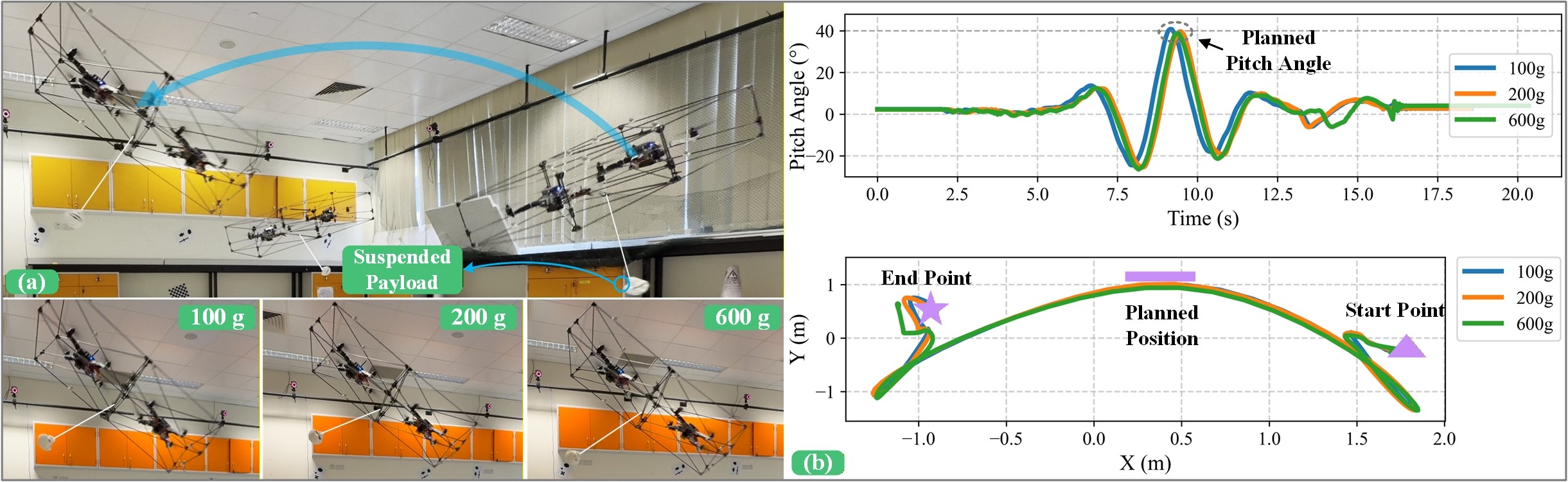}
\vspace{-4mm}
\caption{Agile transport carrying 100 g, 200 g, and 600 g payloads. (a) Pitch angle responses along agile trajectories at $40^\circ$ without retuning the APC parameters, demonstrating robustness to varying unmodeled payload disturbances. (b) Overlaid pitch angle trajectories for different payloads at the planned pitch angle and position, further highlighting robust tracking performance of the agile flight controller.}
\label{fig:real_agile_transport}
\vspace{-6mm}
\end{figure*}
The MARS built from equal-arm quadrotors relies on control allocation for docking, separation, and cooperative flight~\cite{Modquad,ModQuad-Vi,saldana2019design}.
Its angular acceleration capability degrades as the number of units increases~\cite{Modquad}, making accurate trajectory tracking difficult.
As a result, no prior work has demonstrated physical agile tracking experiments using quadrotor-based MARS. We present the first set of real-world demonstrations.
\subsubsection{Circular Trajectory Tracking}

Figs.~\ref{fig:trajectory_tracking}(a) and (b) show circular trajectory tracking of MARS without and with the docking mechanism, respectively.
The composite image captures the entire trajectory along the 1.5 m radius circular path, demonstrating the stable and smooth flight. 
The experimental results confirm that our proposed control framework (model abstraction and abstracted predictive controller) achieves accurate position and attitude tracking throughout the maneuver.


\subsubsection{Collision-free Navigation}
We demonstrate that the MARS can accurately follow ideal collision-free trajectories generated by \cite{huang2025mars-iros}. 
Fig.~\ref{fig:trajectory_tracking}(c) shows the entire flight experiment and plots the desired trajectories together with the actual flight paths.
The results show that the MARS maintains accurate tracking throughout the entire maneuver.
In summary, this experiment demonstrates that an APC controller can directly track planner-generated trajectories for MARS, enabling trajectory planning in the same manner as for conventional aerial robots using only the MARS geometry and model abstraction.

\subsubsection{Agile Flight}
To further evaluate the agile flight capability of the MARS, we test pitch- and roll-angle tracking in Figs.~\ref{fig:agile_trajectory}(a) and (c), and high-speed tracking of a circular trajectory in Fig.~\ref{fig:agile_trajectory}(d). 
Specifically, 
Fig.~\ref{fig:agile_trajectory}(a) illustrates that, using the proposed model abstraction with APC, the $1.2~\text{m}$-long MARS can execute a $3~\text{m}$ agile flight trajectory within a confined space. 
The pitch-angle maneuver can reach $20^\circ$, $30^\circ$, and $40^\circ$, as shown in Fig.~\ref{fig:agile_trajectory}(a) below.
The results in Fig.~\ref{fig:agile_trajectory}(b) show that our control framework not only tracks position accurately but also achieves the desired pitch angles at the intended positions. Notably, our platform ($1200 \times 570~\text{mm}$) is substantially larger than typical agile aerial robots ($160\times160~\text{mm}$ or $100 \times 100~\text{mm}$~\cite{kaufmann2023champion,hanover2024autonomous,wang2025unlocking}), introducing additional difficulties for agile maneuvering. 
Despite this, Fig.~\ref{fig:agile_trajectory}(c) shows roll tracking still reaching $20^\circ$. In Fig.~\ref{fig:agile_trajectory}(d), at speeds above $2~\text{m/s}$, the system tracks a $0.5~\text{m}$-radius circle while sustaining $\sim30^\circ$ roll and pitch. The above demonstrations highlight the powerful tracking capability of the proposed APC framework. Overall, the MARS achieves accurate trajectory tracking and, for the first time, physically demonstrates robust high-speed performance in agile flight.

\subsection{Objct Transport}
\subsubsection{Tracking and Dragging}
In Fig.~\ref{fig:real_transport}(a), the MARS transports a 600 g suspended payload while tracking a 1.5 m radius circle at 2 m/s. Fig.~\ref{fig:real_transport}(c) shows the tracking trajectories. The trajectories for 100 g, 200 g, and 600 g payloads closely overlap the desired trajectory (black dashed line), indicating accurate tracking with different unmodeled payloads. We further test payload-dragging tasks, which are more challenging due to payload-ground contact. The metrics are summarized in Table~\ref{tab:Agile Transport}. The average position error is 0.042 m for tracking and 0.0673 m for dragging. In both cases, the attitude error remains small, demonstrating the robustness of the proposed modeling and controller.
\subsubsection{Sloshing Load}
In Fig.~\ref{fig:transport}, a payload is suspended by a cable and transported along a linear trajectory. Under the proposed APC, the MARS maintains stable flight, achieving mean absolute attitude errors of $1.74^{\circ}$ (pitch) and $0.79^{\circ}$ (roll), with maximum errors of $4.61^{\circ}$ and $3.47^{\circ}$, respectively, and an average position error of $0.17~\mathrm{m}$. Notably, the cable-suspended payload is flexible and influenced by rotor downwash, yet our controller models neither the payload dynamics nor the aerodynamic effects, requires no retuning, and still maintains high-accuracy trajectory tracking. This further highlights the strong robustness of our control framework to unmodeled dynamic disturbances.
\subsubsection{Agile Transport}
Finally, we demonstrate a more challenging task in which MARS carries 100 g, 200 g, and 600 g payloads while tracking an agile flight trajectory, as shown in Fig.~\ref{fig:real_agile_transport}(a). Fig.~\ref{fig:real_agile_transport}(b) shows that the system maintains robust performance at the desired pitch angle of $40^{\circ}$, with a very low average attitude error of $0.79^{\circ}$. Moreover, as reported in Table~\ref{tab:Agile Transport}, the overlapping trajectories achieve an average position error of 0.079 m (MARS size: 1.2 $\times$ 0.6 m), further highlighting the robustness of the proposed framework and its ability to track accurately under unmodeled disturbances from varying payload weights.
\begin{table}[t]
\caption{Comparison of Agile Transport}
\vspace{-3mm}
\label{tab:Agile Transport}
\centering
\scriptsize
\setlength{\tabcolsep}{8pt}
\renewcommand{\arraystretch}{1.15}

\begin{tabular}{@{}lcccc@{}}
\toprule
\makecell{\textbf{Task}} 
& \makecell{\textbf{Payload (kg)}} 
& \makecell{\textbf{Position Error (m)}} 
& \makecell{\textbf{Attitude Error ($^\circ$)}} \\
\midrule
\multirow{4}{*}{Tracking} 
& 100 g & 0.0512 & 2.00  \\
& 200 g & 0.0409 & 2.23  \\
& 600 g & 0.0340 & 1.77  \\
& \textbf{Avg} & \textbf{0.0420} & \textbf{2.00} \\
\midrule
\multirow{4}{*}{Dragging} 
& 100 g & 0.0523 & 2.46  \\
& 200 g & 0.0492 & 2.51  \\
& 600 g & 0.1003 & 2.54  \\
& \textbf{Avg} & \textbf{0.0673} & \textbf{2.50} \\
\midrule
\multirow{1}{*}{Sloshing Load} 
& 530 g & 0.1700 & 1.27  \\
\midrule
\multirow{4}{*}{Agile Transport} 
& 100 g & 0.0696 & 0.85  \\
& 200 g & 0.0613 & 0.27  \\
& 600 g & 0.1061 & 1.26  \\
& \textbf{Avg} & \textbf{0.0790} & \textbf{0.79} \\
\midrule
\textbf{Overall Avg} & -- & \textbf{0.0896} & \textbf{1.64} \\
\bottomrule
\end{tabular}
\vspace{-6mm}
\end{table}
\section{Conclusion and Future Work}
\label{sec:conclusion}
This paper presents MARS-Dragonfly. The biologically inspired docking–locking–separation mechanism enables reliable mid-air docking, detection-free passive locking, and smooth separation within a compact and lightweight design. Beyond hardware, we introduce a unified model abstraction that represents arbitrary MARS configurations as a virtual quadrotor, allowing single-quadrotor controllers to operate across heterogeneous systems. A abstracted predictive controller with real-time allocation ensures actuator feasibility and fully utilizes actuation authority. Extensive simulation and real-world experiments demonstrate low-oscillation docking, smooth separation without large attitude transients, precise trajectory tracking, agile maneuvers, and stable transport of sloshing payloads. These results show that the proposed mechanism, combined with the modeling and control framework, enables MARS to operate as a single quadrotor across configurations and perform complex tasks. Future work will focus on improving the model abstraction and extending it to integrated planning and control.
\bibliographystyle{IEEEtran}
\bibliography{reference}

\newpage

\vfill

\end{document}